\documentclass[twoside,11pt,preprint]{article}
\usepackage[abbrvbib, preprint]{dmlr2e}

\usepackage{macros}
\usepackage{fancyvrb}  

\hypersetup{
    colorlinks=true,
    linkcolor=blue,
    urlcolor=blue,
    citecolor=magenta,      
    }

\usepackage{lastpage}
\dmlrheading{23}{2026}{1-\pageref{LastPage}}{9/23; Revised 10/23}{11/23}{21-0000}{Padidar \& Huang} 
\def\openreview{\url{https://openreview.net/forum?id=XXXX}} 

\ShortHeadings{The BoSS}{}
\firstpageno{1}

\begin{document}
\title{Generative modeling for fusion: a large-scale benchmark of stellarator fast-ion simulations}

\author{\name Misha Padidar\textsuperscript{1}\orcidlink{0000-0002-0710-4377} \email mpadidar@flatironintsitute.org \\
       \name Teresa Huang\textsuperscript{1} \orcidlink{0000-0001-6631-3737}  \\
       \name Mark Goldstein\textsuperscript{1}  \\
       \name Michael Czekanski\textsuperscript{2}  \\
       \name Jeffrey Soules\textsuperscript{1} \\
       \addr \textsuperscript{1}Center for Computational Mathematics, Flatiron Institute\\
       \textsuperscript{2}Department of Statistics and Data Science, Cornell University}

\editor{My editor}

\maketitle

\begin{abstract}
Stellarators offer a promising magnetic confinement approach for achieving controlled nuclear fusion. 
However, stellarator design remains challenging, as evaluating configuration performance requires computationally expensive, high-fidelity simulations.
In particular, key performance metrics derived from guiding center simulations are costly to include directly in the design optimization loop, and are almost exclusively used as diagnostics.
Machine learning has the potential to advance fusion research by enabling rapid generation and evaluation of new stellarator configurations.
Effectively applying ML for fusion demands large-scale datasets that capture both design diversity and high-fidelity simulation outputs. 
In this work, we introduce the Benchmark of Stellarator Simulations (\texttt{BoSS}), a dataset of stellarator designs paired with guiding center simulation results and trajectories. The \texttt{BoSS} is the first large-scale dataset of guiding center simulations, containing simulation data for 180,000 distinct magnetohydrodynamic equilibria spanning a wide range of magnetic geometries and confinement properties. To exemplify the utility of ML to accelerate stellarator design, we formulate and address two sets of modeling tasks: generative modeling for designing stellarators with good particle confinement, and surrogate modeling for predicting stellarator heat deposition patterns and particle trajectories. The tasks reveal the potential impact ML can have in the stellarator field and important opportunities to improve the applicability of ML models in the sciences. Our baselines generate diverse, performant stellarators and provide useful surrogate predictions, and point to long-horizon trajectory forecasting and out-of-distribution generation as key directions for future work. The \texttt{BoSS} can be accessed through \href{https://boss.flatironinstitute.org/}{https://boss.flatironinstitute.org}.
\end{abstract}

\section{Introduction}

Stellarators offer a promising magnetic confinement approach for achieving controlled nuclear fusion \citep{hegna_infinity_2025,swanson_overview_2025,lion_stellaris_2025}. 
The primary goal of stellarator design is to confine a high-temperature plasma (i.e. a quasi-neutral collection of charged particles)  using strong magnetic fields generated by electromagnetic coils. Since the 1980s, the design of stellarators has shifted from relying on analytic approaches, leveraging expert insights and an understanding of fundamental equations, to computational approaches, relying on simulation-based objective functions and numerical optimization routines to improve the performance of designs. 
Computational approaches have led to major advancements in stellarator design \citep{landreman_magnetic_2022,wechsung_precise_2022,goodman_quasi-isodynamic_2024,velasco_piecewise_2024}, 
but a challenge remains to include costly, high-fidelity numerical simulations in the design loop \citep{bindel_direct_2023,jorge_direct_2024,kim_optimization_2024,landreman_bayesian_2026}.


Machine learning (ML) stands to have a significant impact on stellarator design, either by replacing costly high-fidelity simulations in the design loop with fast surrogate models or by bypassing the design loop altogether with generative models of designs \citep{padidar_diffusion_2025,paischer_gyroswin_2025,landreman_how_2025,jenko_accelerating_2025,moreno_investigation_2026,churchill_ai-accelerated_2026}. Surrogates for predicting stellarator properties enable thorough explorations of the design space and serve as rapid diagnostics for stellarator performance.
Moreover, challenging tasks arising from stellarator design can be used to benchmark existing ML methods and accelerate ML for science. However, the success of ML for stellarator design hinges on large-scale datasets that capture both design diversity and high-fidelity metrics. 

A high-fidelity evaluation for stellarator performance is collisionless guiding center (GC) simulations of fast ions \citep{carbajal_alpha-particle_2025,bonofiglo_fast_2025}. These ordinary differential equations predict the trajectories of fast ions, a high-energy product of the fusion reactions. Fast-ion confinement is essential to stellarator performance, for heating the plasma, sustaining the reaction, and preventing damage to plasma-facing components. 

To this end, we introduce the \texttt{BoSS}: a dataset of stellarator designs with fast ion simulation results and trajectories. The \texttt{BoSS} is the first large-scale dataset of fast ion simulation data, containing roughly 180k distinct magnetohydrodynamic (MHD) equilibria -- stellarator designs -- covering a wide range of magnetic geometries, pressure profiles, and confinement capabilities. GC simulations have been run for each stellarator design, with metrics saved as well as initial and final particle states -- the fundamental information needed for analysis and modeling. In addition, detailed particle trajectories have been stored for over 20k stellarators, totalling over 100 million trajectories. Roughly 25k GPU hours and 500k CPU hours were used to generate the data, which occupies 3TB of storage.

To exemplify the utility of the \texttt{BoSS}, we consider two sets of machine learning tasks derived from practical problems in stellarator design: (1) forward modeling—predicting stellarator properties $\yb$ given a stellarator design $\chib$ -- and (2) inverse design -- generating a stellarator design $\chib$ conditional on desired properties $\yb$. 

Within forward modeling, we propose two tasks. Task  (1a) is to predict the heat flux pattern on the plasma boundary due to lost fast ions. As fast ions escape the plasma, they strike plasma-facing components and deposit energy. Predicting the heat flux patterns informs the design of exhaust mechanisms for protecting the device hull. Task (1b) is to predict fast ion trajectories. Trajectories offer an interpretable view of confinement by revealing how and when particles are lost. Due to the chaotic dynamics of fast ions, trajectory prediction serves as a challenging benchmark task for ML trajectory modelling methods.

Within inverse design, we consider the \say{inverse stellarator design} problem from \citep{padidar_diffusion_2025}. Stellarator design is posed as a generative modeling problem with the goal of learning a sample density $p(\chib \, |\, \yb)$ for stellarator designs $\chib$ with desired properties $\yb$, such as low fast ion losses. Progress on this task facilitates more rapid and thorough exploration of the stellarator design space.

For each task, we define concrete \textit{evaluation} protocols as well as diverse baselines, including established methods from stellarator design and various generative modeling approaches (denoising diffusion models, and flow matching). To summarize, the key contributions of our work are the following:
\vspace{-3pt}
\begin{itemize}[leftmargin=*]
    \item The \texttt{BoSS} is the first publicly available large-scale dataset of stellarators with GC simulation data and trajectories. It contains over 180k distinct stellarator equilibria, including many with nonzero pressure profiles, GC simulation data, and over 20k distinct stellarator equilibria with full GC trajectories (over 100M trajectories). The dataset enables training of ML surrogates and fundamental data analysis of fast ion loss mechanisms.
    \item The \texttt{BoSS} includes three practical tasks for informing stellarator design: heat deposition prediction, trajectory generation, and inverse stellarator design. For each task, it provides various ML baselines and introduces standardized evaluation protocols to benchmark progress.
\end{itemize}

\section{Related work}
\label{sec:related_work}
\paragraph{Stellarator datasets} Several large-scale datasets of stellarators have recently been developed: the \texttt{QUASR} repository of quasisymmetric stellarators \citep{giuliani_direct_2024,giuliani_comprehensive_2025}, the Constellaration dataset of quasi-isodynamic stellarators \citep{cadena_constellaration_2025}, and omnigenous stellarators from the Omnigenity database \citep{gaur_exploring_2023}.  
The \texttt{BoSS} builds upon these data sources and extends them in two important ways: introducing pressure and current profile functions to otherwise current-free devices and including high-fidelity metrics from fast ion simulations. Most similar to our work is \citep{landreman_how_2025}, which ran roughly 200,000 gyrokinetic simulations using GX \citep{mandell_gx_2024} for stellarators from the \texttt{QUASR} repository and other sources. Recently, \citet{churchill_ai-accelerated_2026} showcased the use of neural surrogates for gyrokinetic properties to inform stellarator design. While these works have similar computational costs to ours, GC simulations capture an entirely distinct dimension of physics from gyrokinetics.

Other datasets of stellarators include those generated using the near-axis expansion (a fast \textit{approximation} to ideal MHD equilibria) \citep{landreman_mapping_2022,rodriguez_near-axis_2026,curvo_using_2025} and small collections of plasma configurations \citep{kappel_how_2026,buller_family_2025,laia_data-driven_2025}, all of which lack high-fidelity metrics.
Beyond stellarators, datasets of experimental Tokamak scenarios have been developed recently for benchmarking ML methods \citep{spangher2025disruptionbench, rousseau2026tokamarkcomprehensivebenchmarkmast}.

\paragraph{ML benchmarks for shapes, clouds, and particle trajectories}
The \texttt{BoSS} provides a large dataset of toroidal shapes describing magnetic geometries, point clouds representing fast ions lost on the plasma boundary, and particle trajectories from fast ion simulations. Benchmarks for shapes and point clouds such as ModelNet \citep{zhang2025physics}, ShapeNet \citep{chang2015shapenet}, and FAUST \citep{bogo2014faust} focus on vision-centric geometries arising from everyday objects. ShapeBench \citep{fazliani_shapebench_2026} introduces benchmarks for aerodynamic shape optimization, and Stellcoilbench \citep{kaptanoglu_proof--concept_2026} introduces an automated pipeline for stellarator coil optimization.
The \texttt{BoSS} expands this landscape by providing toroidal shapes from stellarator physics. Existing datasets for trajectories include the Anomalous Diffusion (AnDi) Challenge dataset \citep{munoz2021objective} from synthetic anomalous diffusion models and MD17 from the molecular dynamics literature \citep{chmiela_machine_2017,christensen_role_2020}. The \texttt{BoSS} introduces stellarator-based dynamics to the field, featuring unique challenges such as chaos in particle trajectories.

\section{Primer on stellarator physics} \label{sec:primer}
\begin{figure}[tbh!]
    \centering
    \resizebox{\textwidth}{!}{
    \begin{tikzpicture}
        \node (A) at (-0.75,0) {\includegraphics[height=0.175\textheight]{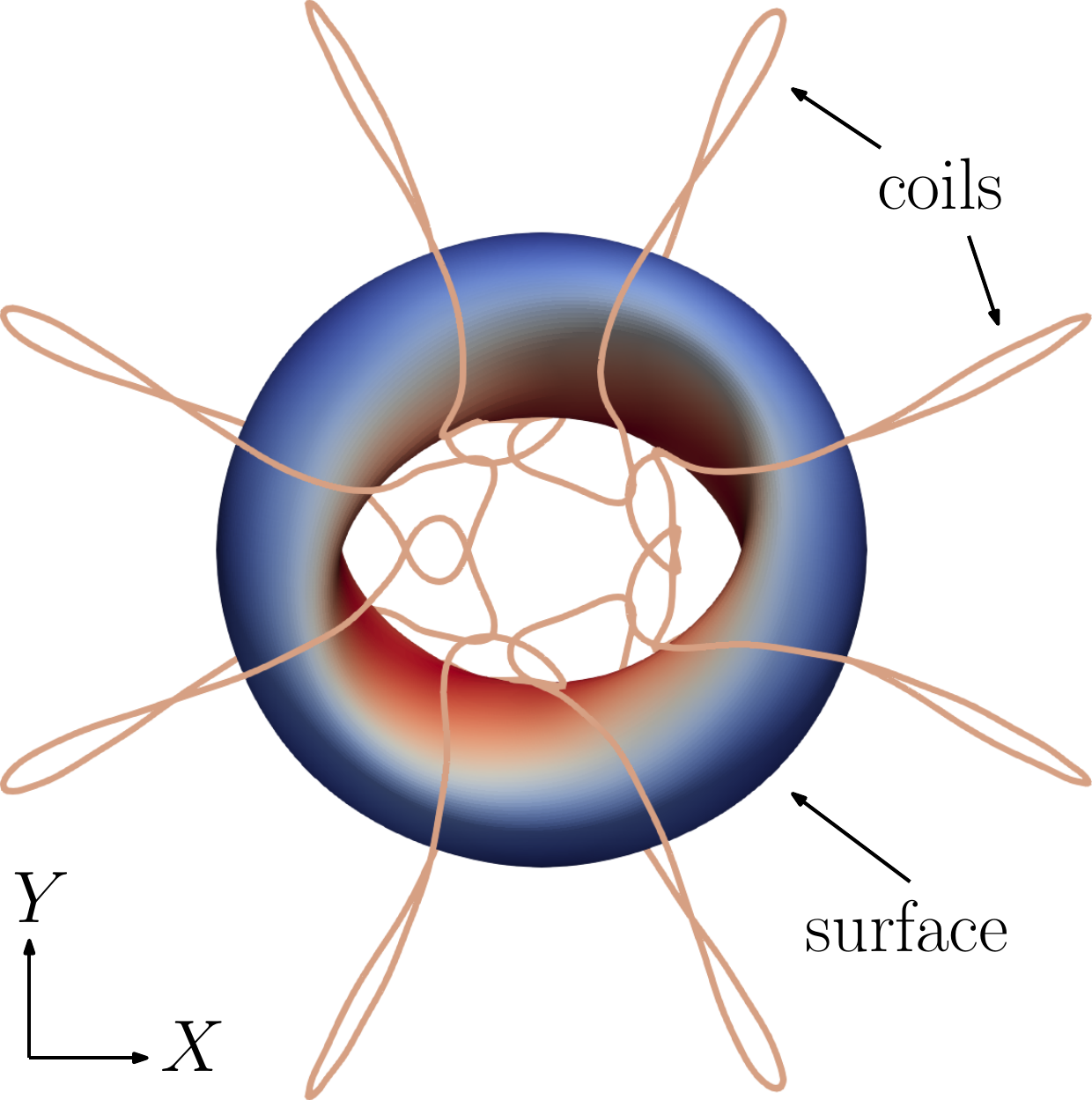}};
        \node (B) at (4.5,0) {\includegraphics[height=0.165\textheight]{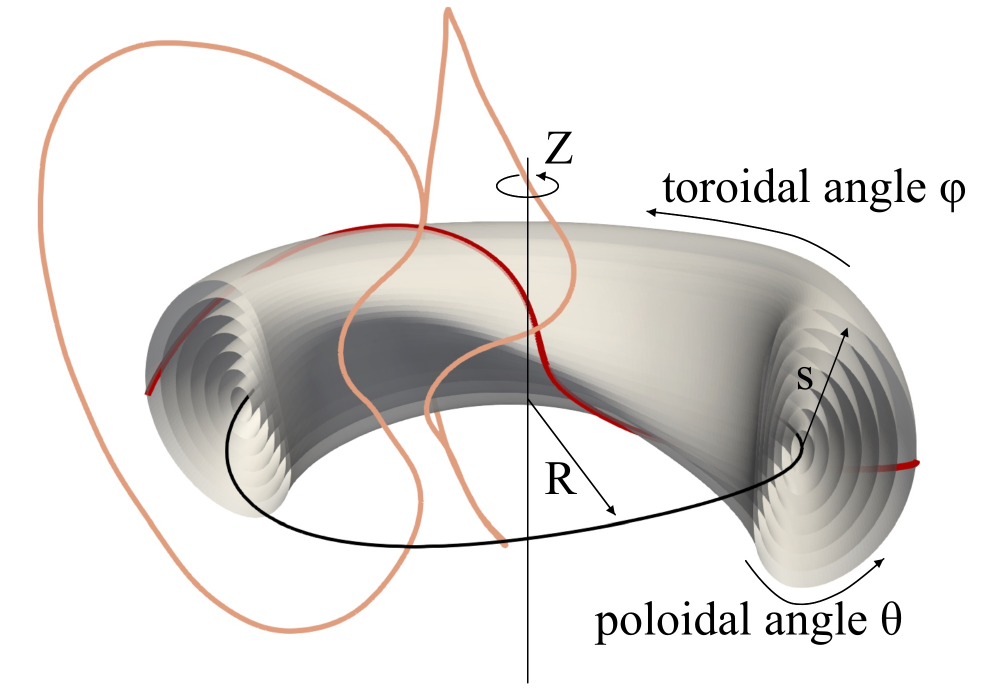}};
        \node (C) at (9.1,0) 
        {\includegraphics[height=0.15\textheight]{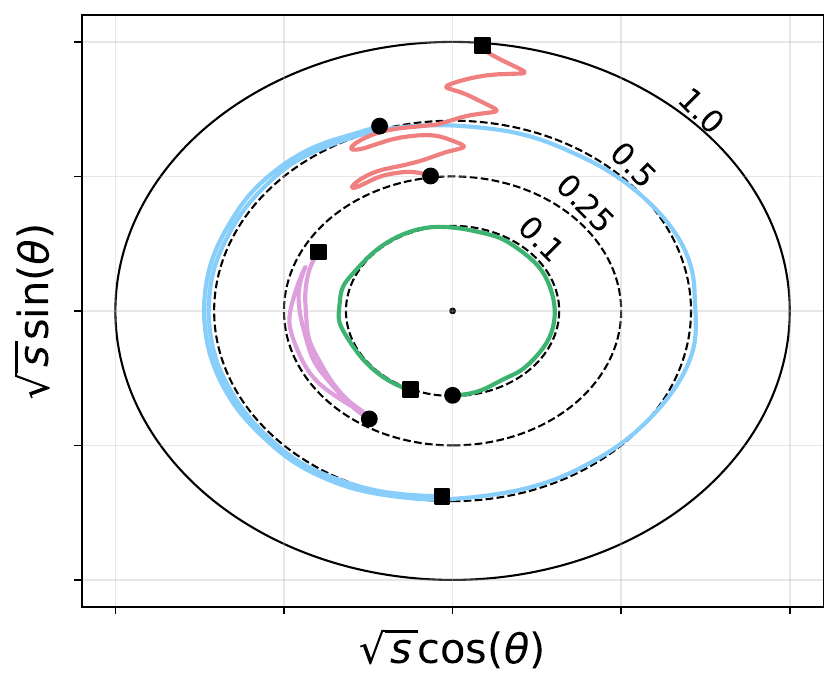}};
        \node (Alabel) at (-1.95, 1.6) {\large A};
        \node (Blabel) at (2., 1.6) {\large B};
        \node (Clabel) at (7.1, 1.6) {\large C};
    \end{tikzpicture}
    }
    \caption{A) The electromagnetic coils and plasma for stellarator-\texttt{0010198} from the \texttt{QUASR} database, as seen from above. The surface is colored by the strength of the magnetic field. B) A slice of the $\nfp=2$-field period device, showing the nested magnetic surfaces, Boozer coordinates $s, \theta, \varphi$, as well as standard cylindrical coordinates $R, Z$. The red curve indicates a hypothetical magnetic field line. C) Segments of four particle trajectories (colored curves) when viewed in a cross-section in Boozer coordinates. The circle marks the start of the trajectory and the square marks the end. The four circles represent four magnetic surfaces with labels $s=0.1, 0.25, 0.5, 1.0$, the outermost being the plasma boundary. One particle intersects with the $s=1$ boundary and is considered lost.}
    \label{fig:stellarator_example}
\end{figure}


Stellarators are a class of toroidal magnetic fields that confine plasmas, ionized gases that are roughly neutral in charge. The trajectories of charged particles that comprise the plasma, ions and electrons, can be manipulated by magnetic fields. The magnetic fields in stellarators are three-dimensional, non-axisymmetric fields \citep{imbert-gerard_introduction_2024}, primarily generated by electromagnetic coils (\Cref{fig:stellarator_example}A). Careful tailoring of the magnetic field not only prevents the plasma from colliding with the device wall but also allows the plasma to support high temperature and density gradients and fusion conditions in the core. Broadly speaking, the goal of stellarator design is to \textit{shape} the magnetic field to confine the plasma and maintain stable equilibrium conditions. Other goals around engineering, heating, and particle exhaust are also important but beyond this introduction; see \citep{imbert-gerard_introduction_2024,boozer_physics_2004} for a thorough discussion.

In equilibrium, the magnetic field in a stellarator, $\Bb$, must satisfy the ideal MHD equilibrium equations:
\begin{equation}
\begin{aligned}\label{eq:idealMHD}
    \mu_0^{-1}(\nabla\times\Bb)\times\Bb &= \nabla p,
    \\
    \nabla \cdot \Bb &= 0,
    \\
    \Bb\cdot \nb &= 0 \quad \text{on} \, \, S(\chib_s),
\end{aligned}
\end{equation}
where $p$ denotes the pressure profile, $\mu_0$ is the permeability of free space, $\nb$ is the outward normal to the plasma boundary surface $S$. The plasma boundary surface, $S$, provides the Neumann boundary condition to the problem. The surface $S$ is typically parameterized by a set of geometric design variables $\chib_s$, such as Fourier coefficients; see \Cref{app:surfaces} for an explicit surface description. 

Stellarator fields are often designed to establish a region of nested magnetic surfaces (\Cref{fig:stellarator_example}B), to which the magnetic field is tangent. Standard cylindrical coordinates $(R, \phi, Z)$ are convenient for describing torii. Nonetheless, the analysis and design of stellarator plasmas conventionally employ a special choice of coordinates that conform to the geometry of nested magnetic surfaces, such as Boozer coordinates $s, \theta, \varphi$. The radial coordinate $s\in[0,1]$ varies outward from the core, identifying magnetic surfaces, while the poloidal and toroidal angles $\theta,\varphi$ label points on a surface (illustrated in \Cref{fig:stellarator_example}B). 
\say{Profile functions} are quantities only depending on $s$, such as the pressure $p(s)$, the current $I(s)$, and rotational transform $\iota(s)$. 

Four fundamental parameters of any stellarator plasma are the number of field periods $n_{\text{fp}}$, the aspect ratio $A$, the rotational transform $\iota$, and the plasma-$\beta$. The number of field periods, $n_{\text{fp}}$, quantifies a discrete rotational symmetry of the device; a rotation of $2\pi/n_{\text{fp}}$ about the $Z$-axis yields an identical configuration. A stellarator with $n_{\text{fp}}$ field periods may be divided into $n_{\text{fp}}$ congruent sections; \Cref{fig:stellarator_example}B shows one of the $\nfp=2$ field periods. The aspect ratio, $A = \Rmajor/\aminor$, is a geometric parameter representing the ratio of the major radius, $\Rmajor$, to the minor radius, $\aminor$, and characterizes the relative compactness of the plasma column. Low aspect ratio configurations are often desirable because they contain substantial plasma volume within a compact device footprint. 
The rotational transform of a magnetic surface $\iota(s)$ represents the mean number of poloidal transits completed by a magnetic field line per single toroidal revolution. \Cref{fig:stellarator_example}B depicts an example field line (shown in red) that approximately completes one poloidal rotation while traversing a single field period, corresponding to $\iota \approx n_{\text{fp}}=2$. Particle confinement is generally enhanced at higher $\iota$ \citep{paul_energetic_2022}. For systems comprising multiple nested surfaces with varying rotational transform values, the quantity $\meaniota$ denotes the value averaged over surfaces. The last quantity, $\beta = \langle\frac{2\mu_0 p }{B^2}\rangle$, is a non-dimensionalized pressure averaged over the plasma volume; $\beta =0\%$ represents a magnetic field with no plasma pressure, while reactor-relevant pressures occur when $\beta$ exceeds $1\%$ or so.

Solving the Ideal MHD equations requires specification of the pressure profile, $p$, the boundary surface $S$, the average magnetic field strength $\langle B \rangle$ (to ensure $\Bb=0$ is not a solution), and a second profile function which can either be the current, $I$, or the rotational transform, $\iota$. Given $(S, p, I, \langle B \rangle)$, the Ideal MHD equations can be solved by numerical simulation code such as \texttt{VMEC}, \texttt{DESC}, \texttt{GVEC}, \texttt{SPEC}, or \texttt{MRx} \citep{hirshman_preconditioned_1991,dudt_desc_2020,hindenlang_computing_2025,hindenlang_gvec_2026,hudson_computation_2012,blickhan_mrx_2025}. \texttt{VMEC} has long been a foundational tool in the stellarator community, while \texttt{DESC} provides a modern JAX implementation, \texttt{GVEC} can simulate a wider class of geometries, and \texttt{SPEC}, \texttt{MRx} have less restrictive physical assumptions.

The magnetic field computed by an MHD solver determines the equations of motion for plasma particles and ultimately the efficiency of a stellarator. Born from the fusion reaction with $3.52\unit{\mega\electronvolt}$ of energy, \emph{fast ions} carry a substantial amount of energy. Without careful design of the magnetic field, fast ions will drift across magnetic surfaces and eventually leave the region of nested flux surfaces. At this point, the particles no longer contribute to the fusion reaction and are considered lost. Sustaining a fusion reaction requires that the energy held within fast ions be confined as long as possible, ideally until the slowing down time, when fast ions slow to the speeds of the thermalized plasma particles (usually $\approx 0.1\unit{\sec}$). 

The behavior of these single particles can be described by an ordinary differential equation. Let the Cartesian position and velocity of a fast ion be $(\xb, \vb)$, and let $B = \|\Bb\|$ denote the magnetic field strength, $\vpar = \vb \cdot \frac{\Bb}{B}$ denote the velocity parallel to the magnetic field, and $v^2 = \|\vb\|^2 = \vpar^2 + \vperp^2$ denote the speed. Ignoring the effect of collisions, the motion of fast ions can be modeled by the \emph{guiding-center equations} (GC) \citep{de_blank_guiding_2000}:
\begin{equation}
    \begin{aligned}\label{eq:guiding_center}
    \dot{\xb} &= \vpar\frac{\Bb}{B} + \vb_D,
    \quad \quad 
    \dot\vpar =\frac{-\mu}{B}\Bb\cdot\nabla B,
\end{aligned}
\end{equation}
where $\vb_D$ captures the drift of particles across field lines and magnetic surfaces. The full formulation in Boozer coordinates can be found in \Cref{app:guiding_center_equations}.
The magnetic moment, $\mu = \frac{\vperp^2}{2B}$, and the kinetic energy, $\frac{1}{2}mv^2$, are both conserved by the guiding center equations, a time autonomous, Hamiltonian dynamical system. While the guiding center equations are a deterministic set of differential equations, the system can experience randomness through the particle initial conditions, and the particle trajectories exhibit chaos. The fast ion birth distribution, $f_0(\xb,\vb)$, is modeled to be proportional to the local fusion reaction rate, which uniformly distributes particles across a flux surface, and isotropically in $\vpar$ \citep{richardson_2019_2019,drevlak_fast_2014,bader_modeling_2021,leviness_energetic_2022}.

Each particle has an associated confinement time $T$ -- the time at which the particle crosses the last closed flux surface, $s=1$ (see \Cref{fig:stellarator_example}C) .
A standard metric for stellarator performance is the loss fraction $\lossfrac = \mathbb{P}(T < \tmax)$ -- the fraction of fast ions lost within some time $\tmax$ -- which is evaluated through GC simulations. A performant stellarator design will confine all particles through the slowing down time; at which point other transport effects dominate fusion performance. Roughly speaking, $10\%$ losses or less, through $\tmax=0.1\unit{\sec}$ is considered performant. Another metric for stellarator performance is energy loss $\energy = \Ebb[3.52e^{-2T/\tmax}]$ -- when $\tmax$ is the slowing down time, this captures the amount of energy lost due to escaping particles \citep{helander_collisional_2005}. The exponential decay captures the mean effect of electron drag. 

While fast ion loss analysis captures only one dimension of a stellarator's performance -- ignoring many objectives such as turbulent effects and MHD stability \citep{bindel_understanding_2023} -- it is a required evaluation benchmark; \emph{a performant stellarator must have good fast ion confinement}. Achieving good fast ion confinement is difficult in practice: estimating $\lossfrac$ requires integrating thousands of chaotic particle trajectories over long time horizons, and the resulting Monte Carlo estimate is noisy and nonsmooth in the design parameters. As a result, designers typically optimize cheaper proxies, such as quasisymmetry \citep{landreman_magnetic_2022}, which correlate only imperfectly with losses, and use GC simulations only to diagnose a finished design.
Having a reliable understanding of which stellarator designs have good fast ion confinement, and a mechanism to generate them frees designers to focus on more challenging aspects of the problem.

\section{Dataset generation}
\label{sec:pipeline}

\begin{figure}
    \centering
    \includegraphics[width=\linewidth]{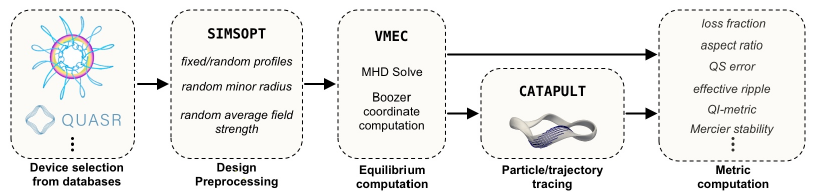}
    \caption{Pipeline for data generation. Devices are selected from candidate datasets such as \texttt{QUASR} and Constellaration, assigned profile functions, and rescaled. \texttt{VMEC} is then used to compute the ideal MHD equilibria before running guiding center simulations with CATAPULT. Metrics are computed from the geometry, magnetic field, and guiding center simulation results.}
    \label{fig:pipeline}
\end{figure}

 The \texttt{BoSS} is a dataset of stellarator designs and simulation-based metrics. Most stellarator designs within the \texttt{BoSS} have been drawn from multiple open-source datasets housing over 1.5 million designs, including the \texttt{QUASR} repository of quasisymmetric stellarators \citep{giuliani_direct_2024}, the Constellaration dataset of quasi-isodynamic stellarators \citep{cadena_constellaration_2025}, a collection of significant stellarators from \citep{kappel_magnetic_2024}, and omnigenous stellarators from the Omnigenity database \citep{gaur_exploring_2023}. To more densely cover typical stellarator parameter ranges, \texttt{QUASR} devices are not added to the \texttt{BoSS} if their aspect ratio exceeds $10$, mean rotational transform is less than $0.1$, or the number of field periods exceeds $6$. To enhance diversity, the \texttt{BoSS} also includes roughly 8,000 rotating ellipse stellarators (stellarators with elliptic cross sections) randomly generated using a procedure similar to \cite{landreman_how_2025}, as well as $21,000$ new devices added using an active learning procedure to improve model performance, described in \Cref{sec:active_learning}. 

The \texttt{BoSS} was constructed through a semi-automated workflow, comprised of three primary steps, as shown in \Cref{fig:pipeline}. The first step selects a stellarator design from a preexisting data source and modifies it by rescaling the size or field strength and including (randomized) $\iota$ or pressure profile functions. Subsequently, the ideal MHD equilibrium is computed along with important properties of the geometry and the magnetic field. Lastly, guiding center simulations are run, confinement data is stored (confinement times with initial and final particle states), and trajectories are optionally saved. The \texttt{BoSS} includes over 180,000 distinct MHD equilibria equipped with GC simulations, forming pairs of stellarator design features, $\chib$, and properties $\yb$. Particle trajectory data in Boozer coordinates is available for over 20,000 devices. 

Pushing one stellarator design through the pipeline can take between $12$ and $55$ minutes with one CPU and one GPU, averaging $15$ when only saving confinement data and closer to $20$ minutes when also saving trajectories; the variation comes from the cost of adaptive timestepping to resolve rapidly varying particle orbits. To conserve GPU resources, the guiding center simulations are the only calculation performed on a GPU (an A100), while other components of the pipeline are CPU-only. Storing device data with confinement data consumes roughly $12\unit{\mega\byte}$ per device and roughly $230\unit{\mega\byte}$ with compressed trajectory data. When compressed in HDF5 format, the database occupies over $3\unit{\tera\byte}$ of storage; the sum of 25,000 GPU hours and over 500,000 CPU hours of compute.

\begin{figure}[tbh!]
    \centering
    \resizebox{\textwidth}{!}{
    \begin{tikzpicture}
        \node (A) at (-0.75,0) {\includegraphics[height=4.5cm]{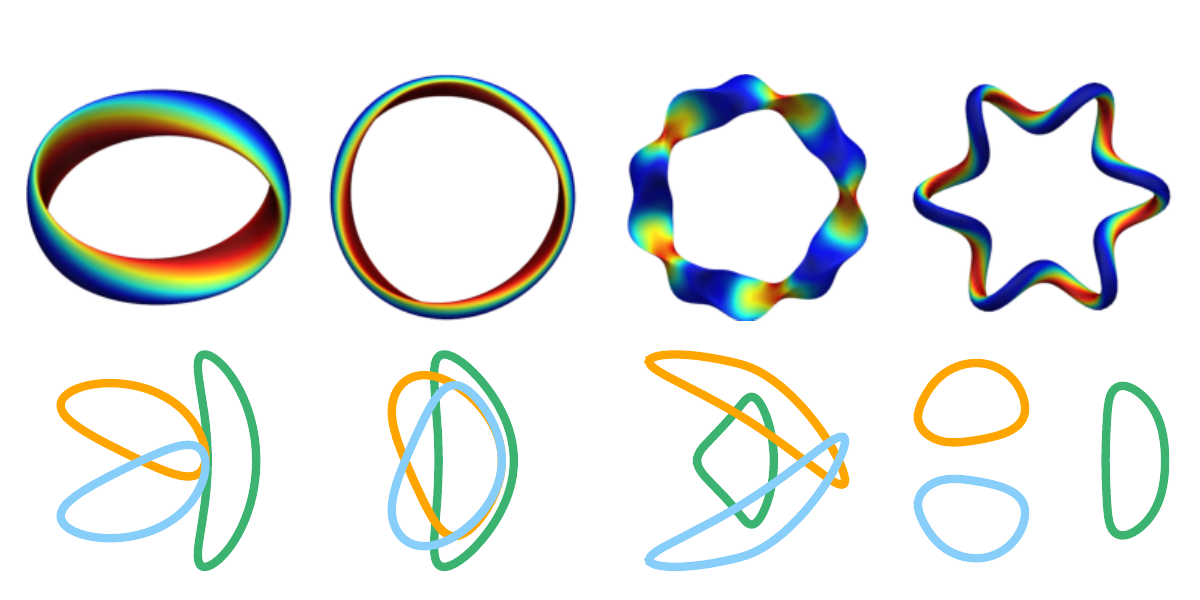}};
        \node (E) at (6,0) {\includegraphics[height=4.5cm]{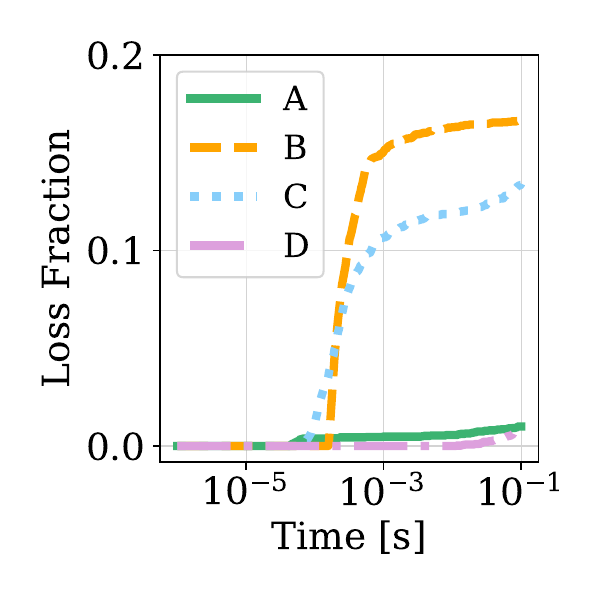}};
        \node (Alabel) at (-5, 1.8) {\large A};
        \node (Alabel) at (-2.7, 1.8) {\large B};
        \node (Alabel) at (-0.5, 1.8) {\large C};
        \node (Alabel) at (1.8, 1.8) {\large D};
        \node (Elabel) at (4., 1.8) {\large E};
    \end{tikzpicture}
    }
    \caption{(A-D) (top) top-down view of four devices taken from the \texttt{BoSS}, colored by the magnetic field strength; (bottom) cross-sections of each device at three fixed values of cylindrical $\phi$. (E) The fraction of particles lost by time $t$ for each device (A, B, C, D).}
    \label{fig:dataset_examples}
\end{figure}

\paragraph{Stellarator parameterization}
Stellarator designs are parameterized through Fourier coefficients $\chib_s$ -- describing the shape of the last closed flux surface $S$ -- the volume average magnetic field strength $\langle B\rangle$, the pressure profile $p(s)$, and either the current profile $I(s)$ or the rotational transform profile $\iota(s)$. The magnetic field is determined from these quantities by solving \Cref{eq:idealMHD}. The boundary shape may exhibit $\nfp$ field period symmetry and often stellarator symmetry: an up-down symmetry in the magnetic geometry, see \citep{imbert-gerard_introduction_2024}. The profiles are saved as a discrete set of values on a grid of $n_s=15$ points, making the vectors $\pb = (p(s_1), \ldots, p(s_{n_s}))$ $\Ib = (I(s_1), \ldots, I(s_{n_s}))$. A stellarator can be represented through the set of variables $\chib = (\chib_s, \nfp, \langle B\rangle, \pb, \Ib)$. In many of our numerical experiments the Fourier coefficients of the surface are replaced with the $R,Z$ coordinates of points on a surface, which can be Fourier transformed to the coefficient representation. \Cref{fig:dataset_examples} shows the boundary shape of four devices taken from the database, and the fraction of particles lost over time in each device.

\paragraph{Device randomization and rescaling}
To build useful models, it is important that a dataset encodes sufficient sensitivities of output quantities $\yb$ with respect to design parameters $\chib$. To capture sensitivities and create heterogeneity, each candidate device drawn from a source database undergoes randomization. The last closed flux surface and edge toroidal flux are rescaled such that the minor radius, $a$, and volume-averaged magnetic field strength, $\langle B\rangle$, have a random practical scale: $a\sim U[0.4, 2.4] \unit{\meter}, \, \langle B\rangle\sim U[4, 10] \unit{\tesla}$, where $U$ denotes the continuous uniform distribution. This range of devices includes reactor scale designs \citep{hegna_infinity_2025,najmabadi_aries-cs_2008} as well as mid-size stellarators \citep{swanson_scoping_2025}. Solving for an ideal MHD equilibrium requires a pressure profile $p(s)$ and either the current profile $I(s)$ or the rotational transform profile $\iota(s)$. The profiles were chosen to be either the default profiles associated with the configuration (from its source database), vacuum profiles ($p=I=0$), a random pair of pressure and rotational transform profiles, or a random pressure profile with zero current ($I=0$). Similar to \citep{landreman_how_2025}, we define the random pressure profiles as non-negative, decreasing quadratics, which ensure the equilibrium has a desired $\beta$ value: $p(s) = p_0 (1 - (1 + c)s + cs^2)$ with $c\sim U[0,1]$ and $p_0 = \frac{\beta \langle B\rangle^2}{\mu_0 (1 - c / 3)}$. The random rotational transform profiles are defined as the original rotational transform profile of the configuration plus a 20\% random linear perturbation: $\iota(s) = \iota_0(s) + l(s)$ where $\iota_0(s)$ is the rotational transform profile of the original design and $l(s) = U_1\iota_0(0) + (U_2\iota_0(1) - U_1\iota_0(0))s$ and $U_1, U_2 \sim U[-0.2, 0.2]$.

\paragraph{Evaluation of MHD equilibria}
A primary goal of the \texttt{BoSS} is to capture the relationships between stellarator designs $\chib$ and fast ion losses. 
To this end, for each device in the database, we store design variables $\chib$, metrics derived from the MHD equilibrium, and metrics derived from guiding center simulations. Computation of MHD equilibria is performed by \texttt{VMEC} \citep{hirshman_steepest_1983,hirshman_preconditioned_1991} through \texttt{SIMSOPT} \citep{landreman_simsopt_2021}. Metrics derived from the equilibrium include the quasisymmetry error, the mirror ratio $\frac{\max B}{\min B}$, geometric quantities such as the aspect ratio, minor radius, plasma volume, and properties of the profiles such as the parallel current $\langle J\cdot \bb\rangle$. We also include stability metrics: the vacuum magnetic well and Mercier stability criterion. We discard devices when \texttt{VMEC} fails to converge because of stringent tolerances, unphysical inputs (LCFS, profiles), or complex magnetic geometries. 

\paragraph{Guiding center tracing}
Guiding center simulations are performed in Boozer coordinates with the \texttt{CATAPULT} code \citep{czekanski2026catapult} through \texttt{FIRM3D} \citep{paul2026firm3d}. \texttt{CATAPULT} is a CUDA accelerated particle tracing code leveraging interpolants of the magnetic field and adaptive timesteppers \citep{dormand_runge-kutta_1986} to accelerate tracing. Generating highly accurate interpolants is a computational bottleneck in the pipeline consuming roughly 6 minutes of compute for an interpolant on a $25\times 25\times 25$ grid in Boozer coordinates. 
To conserve GPU resources, the interpolant construction is performed on CPU so that only the guiding center simulations are performed on a GPU (an A100).
Tracing particles to $\tmax = 0.1\unit{sec}$ takes roughly 5 minutes on average when using both absolute and relative tolerances $5\times 10^{-7}$ for time-stepping. The tracing can take up to $40\unit{\min}$, as the adaptive time-stepper reduces its step size to simulate pathological particle orbits. For the majority of configurations, only initial and final particle states $(s,\theta,\varphi, \vpar)$ are recorded in addition to the confinement times. In this case, $15,000$ particles are traced to $\tmax$; particles that reach $s=0.98$ are considered lost, and their tracing is terminated. 
Devices are discarded if the guiding center simulations fail; numerical issues with adaptive time-steppers near the $s=1$ boundary can infrequently cause failures.

Particles are initialized uniformly over the $s=0.1, 0.25, 0.5$ flux surfaces (5000 particles per surface), and $\vpar \in U[-v, v]$ with $v = \sqrt{2E/m}$ corresponding to an isotropic velocity distribution. Particles represent alpha particles produced by a deuterium-tritium fusion reaction moving with $E = 3.52\unit{\mega\electronvolt}$ of energy and mass, $m = 2m_p + 2m_n$ where $m_p, m_n$ are the proton and neutron masses.
When saving full trajectories, $6000$ particles are randomly traced from the three initial surfaces. Trajectories are traced until $1\unit{\milli\second}$ or until lost, and the particle states are recorded on $1\unit{\micro\second}$ intervals.

\subsection{Expanding the dataset with active learning}
\label{sec:active_learning}

\begin{figure}[hbt!]
    \centering
    \makebox[\textwidth][c]{%
        \includegraphics[width=1.1\textwidth]{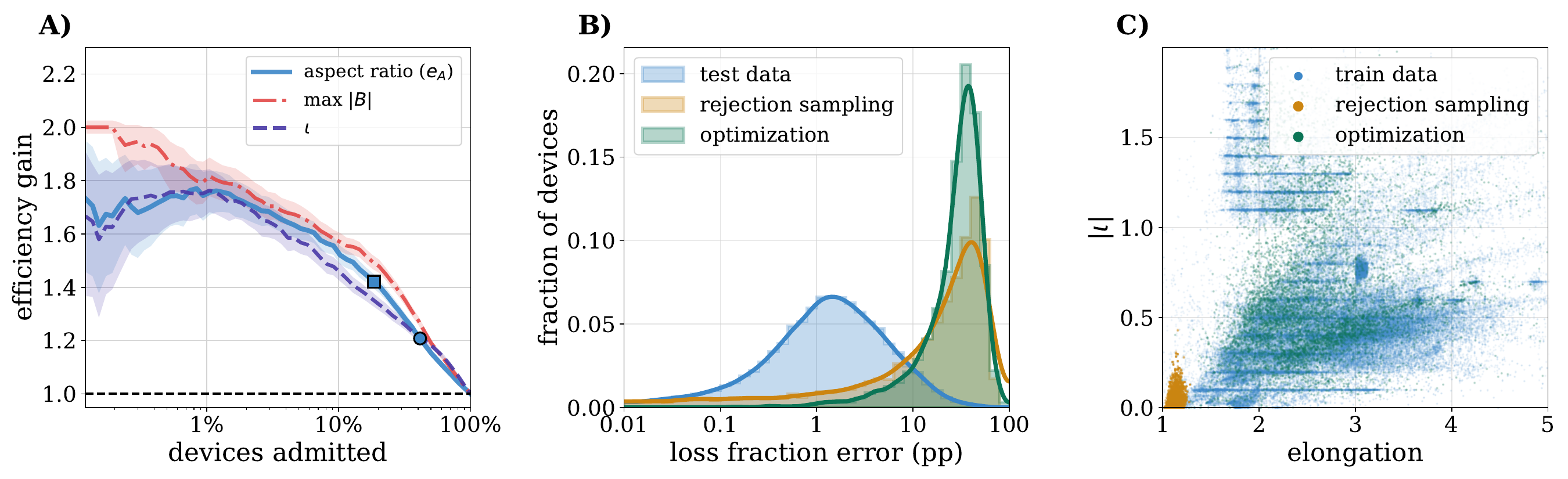}%
    }
    \caption{(A) Efficiency gain of the rejection sampler over uniform random sampling of the the \texttt{BoSS} test set when using three proxy metrics for $e_{\text{loss}}$: $e_A$ and prediction error in $\meaniota, \max B$. Markers show the thresholds used in active learning, $\tau = 0.5\%$ (circle) and $1\%$ (square) at which the efficiency gain is roughly $20\%$ and $40\%$. The band is a 95\% bootstrap confidence interval, and the dashed line marks no gain. (B) Distribution of the loss fraction prediction error $e_{\text{loss}}$ on the test set and on the active learning devices, by collection method. The distribution of high prediction errors highlights the efficiency of the method. Errors below $0.01$ pp ($4.1\%$, $12.5\%$, and $5.4\%$ of each group, respectively) are not shown; these are devices whose predicted and actual loss fractions are both saturated near $0$ or $1$. The units pp indicate percentage points (difference in percentages). (C) The distribution of $|\meaniota|$ and cross-section elongation for training set devices and ative learning devices.}
    \label{fig:active_learning}
\end{figure}

The \texttt{QUASR} and Constellaration datasets, which comprise the bulk of the \texttt{BoSS}, contain only stylized stellarators optimized for good particle confinement and do not cover the broader space of stellarator geometries. ML models trained on this data may have poor predictions in sparsely sampled regions of design space, limiting the applicability of the models. We fill in these gaps using active
learning \citep{kumar_active_2020} to collect over 21,000 additional devices. Active learning aims to collect data points as efficiently as possible so that each new data point maximally improves model predictive performance. Given the high cost of evaluating our pipeline, adding a relatively small, yet impactful, set of data points is a priority.

We focus on collecting new boundary geometries $S$, since the profiles $\pb, \Ib$ can be explored easily by randomization. Following the principles of active learning, new boundaries $\chib_s$ should be collected where the prediction error $e_{\text{loss}}$ of a surrogate model for the loss fraction $m_{\text{loss}}$ is large. However, since evaluating $e_{\text{loss}}$ requires a costly guiding center simulation, we instead use a fast proxy in its place: the relative prediction error, $e_A(\chib_s) = |m_A(\chib_s) - A(\chib_s)|/A(\chib_s)$, of a surrogate for the aspect ratio $m_A$. The key idea is that the aspect ratio and loss fraction models are trained on the same dataset, and since the aspect ratio is a simple geometric quantity and loss fraction depends on particle transport, we expect loss fraction to be at least as difficult to predict as aspect ratio. Thus, large values of $e_A$ should indicate large values of $e_{\text{loss}}$. Aspect ratio can be rapidly computed and differentiated with respect to $\chib_s$, making $e_A$ a practical measure to compute and optimize. 

We also examined using other proxies, such as prediction error in $\bar{\iota}$, but found that $e_A$ was similarly predictive of $e_{\text{loss}}$ and faster to compute. The performance of three proxy metrics -- $e_A$ and the prediction error in $\meaniota, \max B$ -- is compared in \Cref{fig:active_learning}A.
To test the efficiency of each proxy, we rejection sampled points from the the \texttt{BoSS} test set using the proxy as a filter -- we accepted points where the proxy exceeded a tolerance, e.g. $e_A(\chib_s) \geq \tau$ for various values of $\tau$ -- and computed the fraction of this set where $e_{\text{loss}}$ exceeds the test-set median. Since $e_{\text{loss}}$ for a random device exceeds the median with probability $0.5$ by construction, we measure the efficiency gain as $P(e_{\text{loss}} > \text{median} \, |\, e_A \geq \tau) / 0.5$ -- the efficiency gain measures improvement of the rejection sampler over a standard random sampler in finding points with high $e_{\text{loss}}$. \Cref{fig:active_learning}A shows that the rejection sampler achieves a high efficiency: at $\tau = 1\%$, a device that passes the filter is over $40\%$ more likely than a random device to have a large $e_{\text{loss}}$. 

\begin{figure}[tbh!]
    \centering
    \resizebox{\textwidth}{!}{
    \begin{tikzpicture}
        \node (A) at (-0.75,0) {\includegraphics[height=4.5cm]{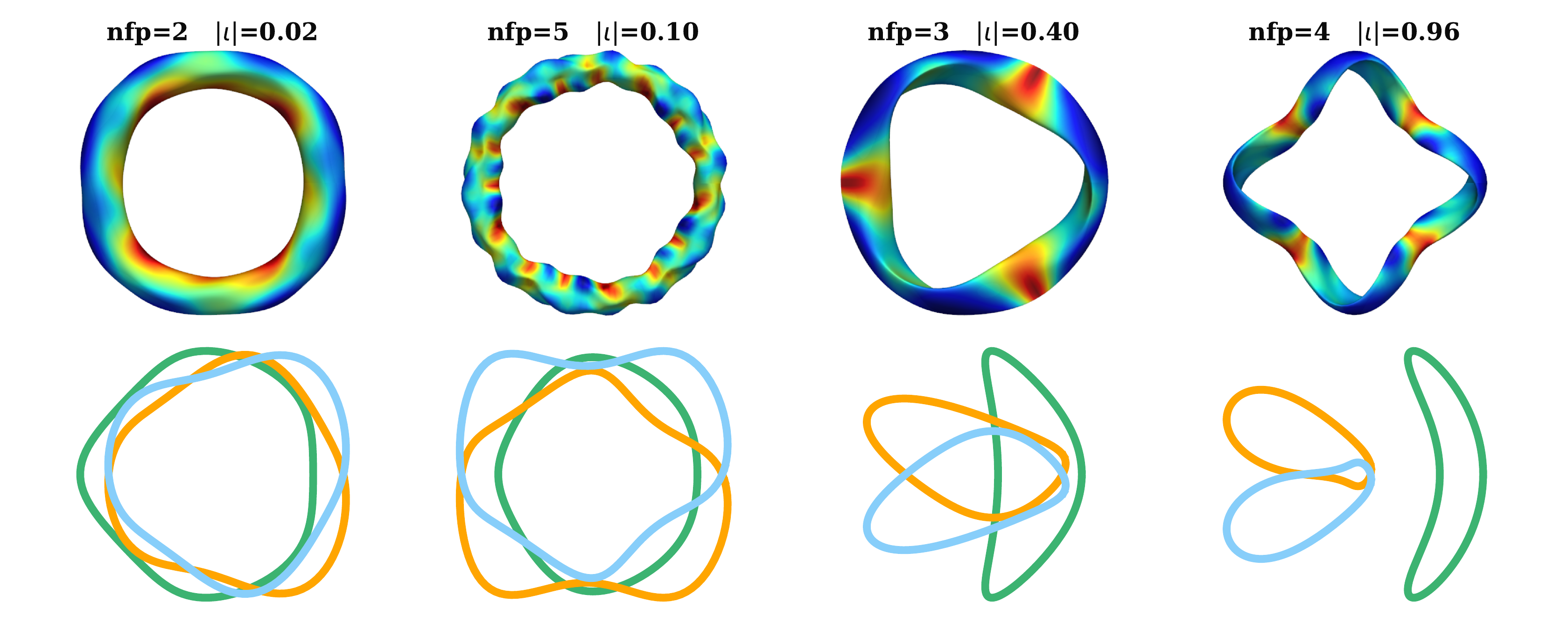}};
        \node (Alabel) at (-6.2, 1.8) {\large A};
        \node (Alabel) at (-3.3, 1.8) {\large B};
        \node (Alabel) at (-0.8, 1.8) {\large C};
        \node (Alabel) at (2.1, 1.8) {\large D};
    \end{tikzpicture}
    }
    \caption{(A-D) (top) top-down view of four devices generated using active learning, colored by the magnetic field strength; (bottom) cross-sections of each device at three fixed values of cylindrical $\phi$. Devices A,B were generated with rejection sampling, while devices C,D were generated with the optimization approach.}
    \label{fig:active_learning_examples}
\end{figure}

New boundary shapes are collected in two ways: with rejection sampling and with optimization. Both approaches are described in detail in \Cref{app:active_learning}. The rejection sampling approach samples boundaries using the random sampler from \Cref{app:surfaces} and accepts them if $e_A \geq \tau$, with $\tau = 0.5\%$ or $1\%$, and if \texttt{VMEC} converges with vacuum profiles. The optimization approach seeks devices with a combination of low predicted loss fraction and large prediction error, $e_A$, at various sizes and aspect ratios. The objective favors designs predicted to perform well but where the loss fraction model may be inaccurate. Both generation methods have distinct advantages: the optimization method can find devices with higher prediction error, and the rejection sampler has a relatively low computational cost.

\Cref{fig:active_learning}B shows that $e_{\text{loss}}$ is significantly higher for devices from the active learning data than for devices from the test set, demonstrating that $e_A$ served as a good proxy for $e_\text{loss}$. As expected, the optimization finds the highest density of points with high loss fraction prediction error. \Cref{fig:active_learning}C shows the distribution of $|\meaniota|$ and cross-section elongation of the sampled devices. Defined in \Cref{app:active_learning}, the elongation of a constant-$\phi$ cross section is the ratio of the extents along the two principal variation directions -- this equals 1 for a circle and $b/a$ for an ellipse with semi-axes $b \ge a$, at any orientation. Due to an intrinsic bias in the random sampler, rejection sampling densely samples devices with low $|\meaniota|$ and low elongation. The optimization method, on the other hand, samples devices over a broad range. Four devices produced by the active learning procedure are shown in \Cref{fig:active_learning_examples}: A,B are generated with rejection sampling, while C,D are generated with optimization. The active learning devices are included in both the train and test sets of the \texttt{BoSS}.

\section{Train and test set statistics}
\label{sec:dataset_stats}
\begin{figure}[tbh!]
    \centering
    \includegraphics[width=\linewidth]{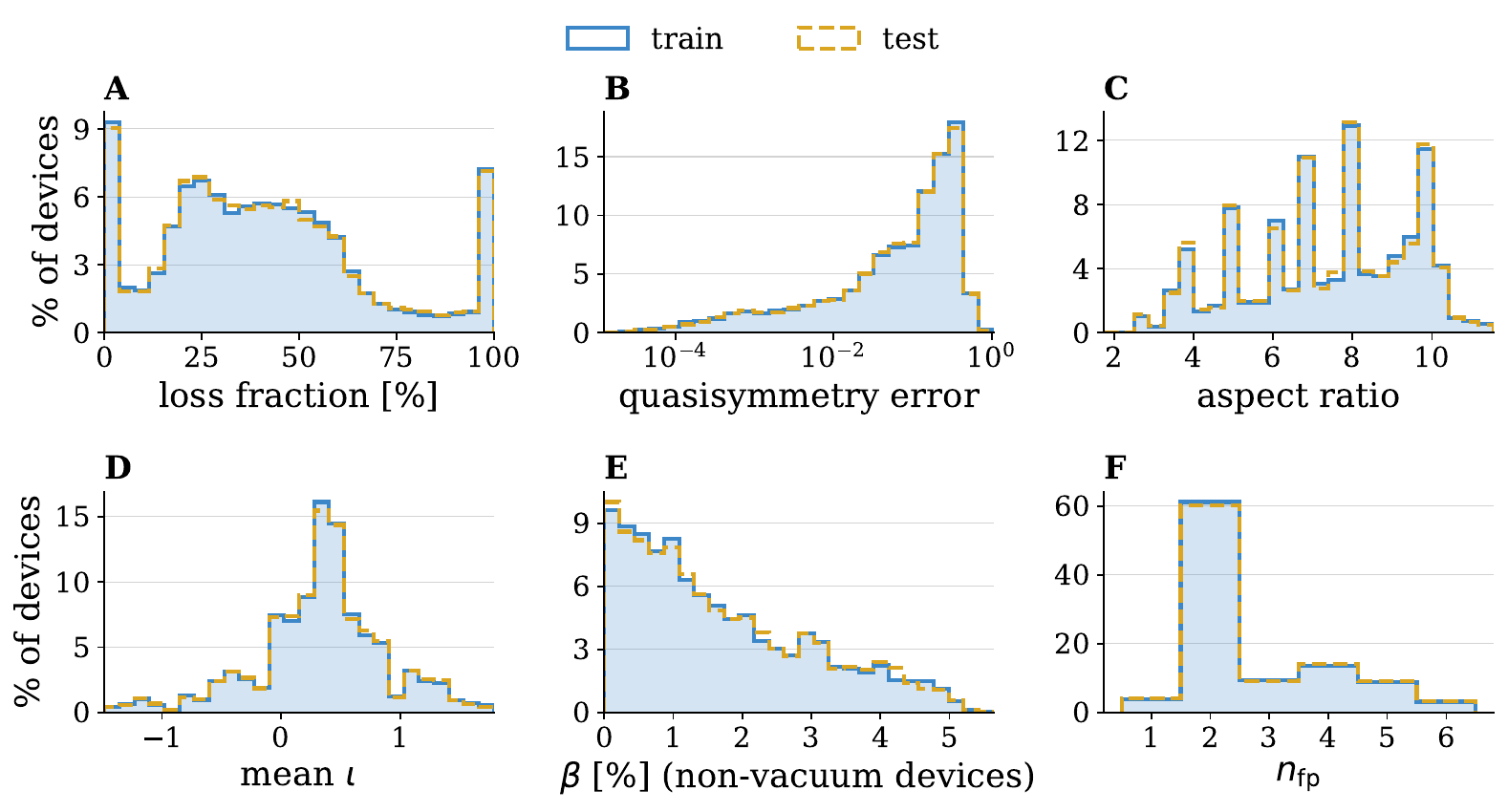}
    \caption{Basic properties for devices in the train and test sets: A) loss fraction for devices born on $s=0.1$; B) the square root of the quasisymmetry error; C) the aspect ratio; D) the mean rotational transform; E) the plasma-$\beta$ for non-vacuum devices; F) and the number of field periods.}
    \label{fig:dataset_statistics}
\end{figure}

The 180k stellarators in the \texttt{BoSS} are divided into train and test splits consisting of roughly 155k points and 25k points, respectively. The test set is selected as a uniform sample from the entire set, followed by a filtering procedure that ensures devices with the same surface geometry but different scales or profiles do not appear in both train and test sets. \Cref{fig:dataset_statistics} shows basic properties of devices in the train and test sets, demonstrating their distributional similarity: (A) loss fraction for devices born on $s=0.1$; (B) the square root of the quasisymmetry error (sum of squared symmetry breaking harmonics normalized by the sum of squared quasisymmetric harmonics); (C) the aspect ratio; (D) the mean rotational transform; (E) the plasma-$\beta$ for non-vacuum devices; and (F) the number of field periods. \Cref{fig:dataset_statistics}B shows that a large fraction of devices have significant deviation from quasisymmetry ($>1\%$), despite many devices being drawn from \texttt{QUASR}, where most devices are near quasisymmetry. \Cref{fig:dataset_statistics}E shows the range of plasma-$\beta$ values in the dataset, reaching reactor-grade values ($>1\%$). Roughly $35\%$ of the dataset has non-zero plasma pressure. 

The \texttt{BoSS} website (\href{https://boss.flatironinstitute.org/}{https://boss.flatironinstitute.org}) provides access to a data visualization tool that facilitates filtering the entire dataset by desired quantities (e.g. $\nfp, A, \beta$, etc.) and also viewing an individual device's geometry, profiles, and fast ion loss curves. The associated \texttt{pip} package \texttt{boss-bench} simplifies the process of opening files, slicing data, computing benchmark metrics, and making plots.

\section{Benchmarks and evaluation}
\label{sec:experiments}

We now consider two sets of machine learning tasks derived from practical problems in stellarator design: (1) inverse design -- generating a stellarator design $\chib$ conditional on desired properties $\yb$ -- and (2) forward modeling -- predicting stellarator properties $\yb$ given a stellarator design $\chib$. Within inverse design, we consider the \say{inverse stellarator design} problem from \citep{padidar_diffusion_2025,curvo_using_2025}, learning a conditional density $p(\chib \, |\, \yb)$ over designs $\chib$ with desired properties $\yb$, such as low fast ion losses. The inverse design bypasses the need for an expensive stellarator optimization loop. 
Within forward modeling, (2a) predicts the heat deposition pattern of lost fast ions on the plasma boundary, which informs the design of plasma-facing components. (2b) generates the fast ion trajectories themselves, acting as a general-purpose surrogate for the GC integrator, from which any confinement metric can be derived.

\subsection{Inverse stellarator design}
\label{sec:inverse_design}

Stellarator design is typically framed as an optimization problem: minimize an objective function over the space of stellarators with design variables $\chib$ to find a design with properties $\yb$. Inverse stellarator design flips this paradigm and learns a probability density $p(\chib \, |\, \yb)$ or an equivalent sampling mechanism, which can produce stellarator design variables $\chib$ for stellarators with properties $\yb$. This approach has two appealing properties. First, it can be seen as a type of amortization: pay up-front to train the model, but then sample new devices at a cost lesser than running new optimizations. Second, generative models are known to generalize; prompting the model with $\yb$ not exactly seen in the data can produce devices that blend aspects of dataset devices but that are also new. As shown in \citep{padidar_diffusion_2025,curvo_using_2025}, the inverse problem can be solved with denoising diffusion models or mixture density networks as an alternative to standard numerical optimization approaches.

We solve the problem from \citep{bindel_direct_2023} seeking stellarators with low fast ion losses and bounded mirror ratio $\rmirror = \frac{\max_{s,\theta,\varphi}{B(s,\theta,\varphi)}}{\min_{s,\theta,\varphi}{B(s,\theta,\varphi)}}$; the constraint prevents solutions with unphysically high variation in the magnetic field. Solving this inverse problem with ML models is particularly advantageous, since evaluating the loss fraction is costly and does not admit derivatives, making traditional optimization methods cumbersome \citep{bindel_direct_2023}. The methods should generate designs with a loss fraction less than $10\%$, a mirror ratio less than $\rmirror^*$, and aspect ratio equal to $A^*$. Since the remaining descriptors of a stellarator equilibrium $\cb = (\pb, \Ib,\nfp, \aminor,\langle B\rangle)$ are free to be selected, only the surface $\chib_s$ must be designed.

\paragraph{Model baselines}
\textit{Diffusion} We train a conditional diffusion model with flow matching \citep{lipman_flow_2023, albergo_stochastic_2023,liu_flow_2023} using a 2D UNet velocity field to generate boundary surfaces. The boundary surfaces are represented as a two-channel image $(R, Z) \in \R^{2 \times 41 \times 41}$ of cylindrical
coordinates over one field period. The model is conditioned on $\yb = (\nfp, \lossfrac, \psiedge, \aminor, A, \Bmin, \rmirror, \pb, \iotab, n_{\text{curr}})$: the number of field periods, the loss fraction, the edge toroidal flux, the minor radius, the aspect ratio, the minimum field strength, the mirror ratio, the pressure and rotational-transform profiles on 15 points in $s$, and a boolean that determines if the current profile is given to \texttt{VMEC} or the rotational transform profile. $\langle B\rangle$, $\Rmajor$, $B_{\max}$ and the current profile are determined by $\yb$ and are not conditioned on. When generating new designs for the results in this section, $\yb$ is sampled from held-out test devices whose own loss fraction is below $10\%$; this ensures the conditions don't preclude the possibility of generating devices with low loss fraction. Similarly, the target mirror ratio and aspect ratio, $\rmirror^*, A^*$, used for computing metrics are the conditioning values.
Architecture, conditioning, and training details are given in \Cref{app:diffusion}.

\textit{Surrogate optimization (SurrogateOpt).}
A second approach to solving the inverse design problem is through numerical optimization, with neural surrogates substituted for quantities computed from particle tracing. The SurrogateOpt baseline generates designs by solving,
\begin{align}
    \label{eq:opt_problem}
    \min_{\chib} J(\chib) = m_{\text{energy}}(\chib)
             + J_{\text{aspect}}(\chib) + J_{\text{minor}}(\chib)
             + J_{\text{mirror}}(\chib),
    \end{align}
where $m_{\text{energy}}$ is a neural surrogate prediction of the normalized energy loss at $t_{\max} = 0.1\unit{\sec}$ for particles launched from $s = 0.1$, and the remaining objectives defined in \Cref{eqn:J_aspect}, \Cref{eqn:J_mirror} encourage the aspect ratio, minor radius, and mirror ratio to achieve targeted values and can be computed exactly by running \texttt{VMEC}. Gradient-based optimization is used, and only the surface degrees of freedom are varied. $\beta$ is fixed to $0\%,3\%$, the mirror ratio target is $\rmirror^*=1.5$, and the aspect ratio target is drawn from $A^*\sim U[3.5,10]$. Solutions produced by the surrogate optimization are discarded if the predicted loss fraction is greater than $10\%$. Details on the surrogate modeling and solving the optimization problem are given in \Cref{app:loss_optimization}. 

\paragraph{Model data}
All methods are trained on the \texttt{BoSS} training split, with the loss fraction defined by particles born on $s = 0.1$ and traced to $t_{\max} = 0.1\unit{\sec}$. All designs are evaluated at the ARIES-CS scale ($\aminor = 1.7\unit{\meter}$, $\langle B \rangle =
5.86\unit{\tesla}$) by re-running the dataset pipeline (\Cref{sec:pipeline}): fixed-boundary \texttt{VMEC} followed by
guiding-center tracing of 5{,}000 alpha particles born on $s = 0.1$. The ARIES-CS scale is a standard device scaling for benchmarking fast ion confinement. Equilibria are computed either in vacuum ($\beta = 0$) or
at finite beta ($\beta > 0$).

\paragraph{Evaluation metrics}
Metrics are computed over designs whose equilibrium and tracing are completed successfully.
\begin{enumerate}
  \item \textit{Low-loss rate.} The fraction of designs whose traced loss fraction is below the target value, $P(\lossfrac \le 10\%)$. We also report $P(\lossfrac \le 15\%)$ to capture near-misses that local refinement could likely bring under the threshold.
  \item \textit{Mirror constraint.} The fraction of designs satisfying the mirror ratio constraint $P(\rmirror \le \rmirror^*)$, and its relaxation $P(\rmirror \le 1.1\rmirror^*)$, reported as \textit{Mirror constraint}-$1.1\rmirror^*$. The diffusion model treats $\rmirror^*$ as a target value rather than a bound, so a design that matches its target to within a percent can fall on either side of it; the relaxed criterion gives a fairer comparison.
  \item \textit{Aspect ratio error.} The median relative error between the aspect ratio and its target value, $|A - A^*|/A^*$. 
\end{enumerate}

\begin{table}[tbh!]
\centering
\caption{Inverse design at ARIES-CS scale, over successfully evaluated designs; metrics are defined in \Cref{sec:inverse_design}. \textbf{Bold} = winner per row. Diffusion is conditioned on the property vectors of held-out test devices whose loss fraction is $\le 10\%$ and evaluated after rescaling to $\aminor = 1.7\unit{\meter}$, $\langle B\rangle = 5.86\unit{\tesla}$; its $\rmirror^*$ and $A^*$ are the conditioning values of each source device, whereas SurrogateOpt uses $\rmirror^* = 1.5$ and $A^* \sim U[3.5, 10]$.}
\label{tab:inverse_design_results}
\begin{tabular}{lcc}
\toprule
Metric & Diffusion & SurrogateOpt \\
\midrule
\textit{Vacuum} ($\beta = 0$) & $N = 417$ & $N = 97$ \\
Low-loss rate-$10\%$ [\%] $\uparrow$ & 43.6 & \textbf{52.6} \\
Low-loss rate-$15\%$ [\%] $\uparrow$ & 48.4 & \textbf{64.9} \\
Mirror constraint [\%] $\uparrow$ & 10.1 & \textbf{20.6} \\
Mirror constraint-$1.1\rmirror^*$ [\%] $\uparrow$ & \textbf{97.4} & 41.2 \\
Aspect error (median) [\%] $\downarrow$ & \textbf{0.3} & 3.7 \\
\midrule
\textit{Finite beta} ($\beta > 0$) & $N = 169$ & $N = 109$ \\
Low-loss rate-$10\%$ [\%] $\uparrow$ & \textbf{60.9} & 37.6 \\
Low-loss rate-$15\%$ [\%] $\uparrow$ & \textbf{71.6} & 50.5 \\
Mirror constraint [\%] $\uparrow$ & 20.7 & \textbf{29.4} \\
Mirror constraint-$1.1\rmirror^*$ [\%] $\uparrow$ & \textbf{98.8} & 45.0 \\
Aspect error (median) [\%] $\downarrow$ & \textbf{0.4} & 3.6 \\
\bottomrule
\end{tabular}
\end{table}

\paragraph{Results}
Diffusion and SurrogateOpt both create novel designs
with good fast-ion confinement. \Cref{tab:inverse_design_results} show the headline performance metrics. Both methods perform similarly in terms of the low-loss rate, producing designs with low losses roughly 40\% to 60\% of the time;
SurrogateOpt outperforms Diffusion when generating designs with $\beta=0$ (vacuum), while Diffusion outperforms SurrogateOpt when generating devices with $\beta >0$ (non-vacuum). For comparison, $10.8\%$ of randomly drawn test devices have less than $10\%$ loss fraction when rescaled to the ARIES-CS scale ($12\%$ at their own scale).


The low-loss rate for Diffusion depends strongly on the conditioning vector: when using conditions drawn from a \textit{random} test device and 
only overwriting the loss target to $10\%$, the low-loss rate at ARIES-CS scale is $6.7\%$ ($7.2\%$ over all loss targets between $2\%$ and $20\%$);
 whereas using conditions from random test devices with less than $10\%$ losses, improves the low loss rate to the values in the table (\Cref{app:diffusion}).
The modest performance of Diffusion in terms of the loss rate highlights an opportunity for post-training: online or offline reinforcement learning could improve the probability of sampling devices with low loss fraction. 

\begin{figure}[tbh!]
    \centering
    \includegraphics[width=\linewidth]{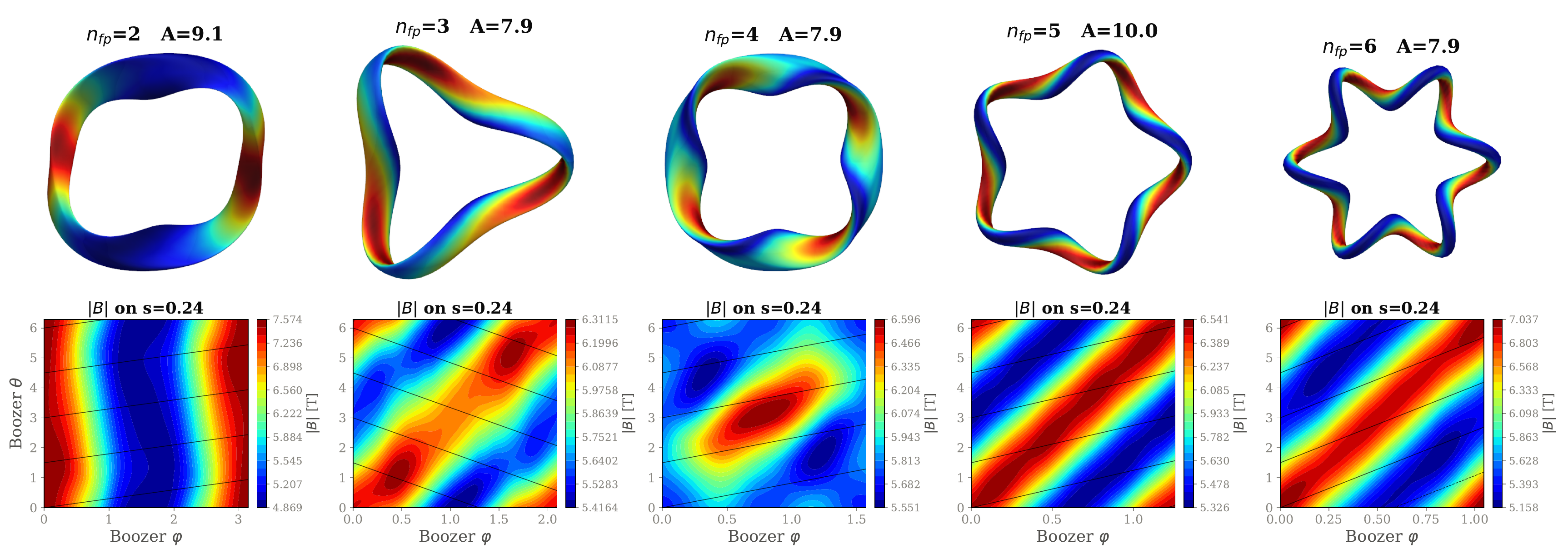}
    \caption{Five designs generated by Diffusion that have $< 10\%$ losses ($5.8\%,4.5\%,4.9\%,0.0\%,0.7\%$ from left to right), mirror-ratio less than $1.65$, and are far from the training set (normalized distance $\ge 0.17$ minor radii to its nearest training boundary, \Cref{app:surfaces}). 
    The $\nfp = 3$ device is a finite-pressure equilibrium, the others are vacuum.
    Top: boundary surface at the ARIES-CS scale coloured by $|\Bb|$. 
    Bottom: $|\Bb|$ in Boozer coordinates on the $s = 0.24$ surface.
    The $\nfp=3,5,6$ devices display a helical symmetries, while the $\nfp=2$ device displays poloidal variation.}
    \label{fig:diffusion_designs}
\end{figure}

Five designs generated by Diffusion are shown in \Cref{fig:diffusion_designs} (top row) above the contours of $B$ in Boozer coordinates on the $s=0.25$ surface. The devices have less than $10\%$ losses, mirror ratio less than $1.65$, and are measurably distinct from devices in the training set. The $B$ contours of the $\nfp=4$ device display piecewise omnigenity \citep{velasco_piecewise_2024}, while the $\nfp=5,6$ device display quasihelical symmetry \citep{landreman_magnetic_2022}. Additional results, including the distance of generated designs to the training set---supporting the novelty of the devices---are given in \Cref{app:diffusion} and \Cref{app:loss_optimization}.

Diffusion matches the aspect ratio constraint and relaxed mirror ratio constraint (Mirror constraint-1.1$\rmirror^*$) tightly, outperforming SurrogateOpt. Diffusion seldom satisfies the exact mirror constraint since it treats $\rmirror^*$ as a target rather than a bound. 
However, since the median overshoot is only $1.3\%$ of $\rmirror^*$, the samples almost always satisfy the relaxed constraint.
 
The solutions produced by SurrogateOpt have relatively large constraint violation because during the surrogate optimization the penalties on the mirror ratio and aspect ratio were left at small values to favor finding low-loss fraction solutions; the constraint violation could be driven to zero by using constrained optimization methods.

The methods also differ in computational cost. Diffusion has an upfront training cost of roughly 80 hours on four A100 GPUs and a sub-second sampling cost on a GPU. On the other hand, SurrogateOpt has a small upfront training cost for building surrogates ( $\sim2$ A100 GPU hours), but a large optimization cost: roughly $82$ CPU cores for $\sim 1$ hour, to evaluate finite-difference gradients through \texttt{VMEC}. The high optimization cost of SurrogateOpt could be mitigated by using a differentiable solver such as \texttt{VMEX}, \texttt{DESC}, or \texttt{MrX}. Due to the overall computational cost of SurrogateOpt,
only 206 solutions were produced with predicted loss fraction $<10\%$.  Evaluating a design with \texttt{VMEC} and particle tracing costs the same for both.

Practically, Diffusion is much easier to use than SurrogateOpt. A well documented problem with SurrogateOpt \citep{fazliani_shapebench_2026}, is that the optimizer may find undesirable local minima of the modeled objective or regions of space where the model generalizes poorly. Diffusion does not suffer from this drawback, by sampling regions where the models are fairly confident. Despite having this drawback, SurrogateOpt does find a host of good solutions that are distinctly different from the training set and have satisfactory loss fraction and mirror ratio (\Cref{app:loss_optimization}).

\subsection{Forward stellarator design}

Determining the fast ion confinement properties of stellarator requires \emph{simulating particle trajectories} through the device. This is an expensive simulation, and can usually only be used to evaluate a stellarator design and not design a stellarator. Our goal is thus to \textit{learn} surrogates that can predict fast ion loss statistics in stellarators. This section will introduce two tasks: prediction of heat deposition patterns on the plasma boundary and prediction of chaotic trajectories of particles.

\subsubsection{Heat flux prediction}
As fast ions escape the plasma, they collide with plasma-facing components (walls, sensors, etc.) depositing energy and degrading the components. The plasma-facing components cannot withstand arbitrary heat loads, and so the device must be designed to control the fast ion heat flux. 
To better protect the device, it is crucial to have rapid predictors of the fast ion heat-deposition pattern, which can be formulated as an ML task as follows: given the device design $\chib$ and the number of fast ions $n$, predict the lost fraction $\lossfrac$ jointly with the positions and energies of lost ions $\{(\xb_i, E_i) \}_{i=1}^{n_{\text{lost}}}$ as they cross the plasma boundary ($s=1$), where $n_{\text{lost}} = n \, \lossfrac$. From the distribution of the \textit{lost particle} positions and energies on the plasma boundary, $p(\xb, E)$, we can estimate the heat flux as $Q_p(\xb) = \lossfrac\int E\,p(\xb,E)\,dE \, \, [\unit{\mega\electronvolt}/\unit{\meter^2}]$. A downstream task could use these heat flux estimates to inform the design of divertors.

\paragraph{Model baselines} We frame the heat deposition prediction as a generative modelling task, due to its probabilistic nature (e.g., randomness from the particle initial positions). To this end, we use Conditional Flow Matching with two different backbones for the velocity field prediction \emph{conditional on $\hat{n}_{\text{lost}}$} (an estimate of $n_{\text{lost}}$): (a) per-point baseline that operates on each fast ion independently (labeled IID model), and (b) PointNet baseline which adds an inter-point feature aggregation module to (a). Both backbones share the same condition encoder for $\chib$, the time embedding, and the loss function, $\ell = \ell_{\text{CFM}} + \lambda \ell_{\text{loss}}$, combining both the standard CFM loss and the auxiliary loss to predict the loss fraction. At inference time, both variants encode $\chib$, predict $\hat{n}_{\text{lost}}$, and then generate $\hat{n}_{\text{lost}}$ lost ion positions and energies with flow matching. The full model architectures are described in \Cref{app:heat_prediction}.

\paragraph{Model data}
We set the design parameter $\chib = (\hat{\chib}_s, \cb, s_{\text{init}})$ where $\hat{\chib}_s\in \R^{3 \times 41 \times 41}$ represents Cartesian grid points on the boundary surface, the conditions $\cb = (n_{\text{fp}}, \aminor, \langle B \rangle, \mathbf{\frac{dp}{ds}}, \iotab)$ are described in \Cref{sec:primer,sec:pipeline}, and $s_{\text{init}} \in \{0.1, 0.25, 0.5 \}$ represents the inner surface on which the fast ion is born (see \Cref{fig:stellarator_example}C). Conditioned on $\chib$, we then filter out lost ions where confinement time $T < t_{\max}$, and construct the prediction target $\yb = (\lossfrac, \{(\xb_i, E_i) \}_{i=1}^{n_{\text{lost}}})$, where the particle positions, $\xb$, are represented by the Boozer coordinates $(\theta, \varphi)$. The positions $\theta_i, \varphi_i$ are mapped to the torus by the \textit{angular embedding}
$(\cos\theta_i,\sin\theta_i, \cos\varphi_i,\sin\varphi_i)\in\mathbb{R}^4$ before being fed to the predictive models.

\paragraph{Evaluation metrics}
\begin{enumerate}
    \item \textit{Energy Distance (ED).}
A normalized statistical distance between point clouds $X \in \R^{n\times d}$ (predicted) and
$Y \in \R^{n'\times d}$ (true) with the Euclidean metric,
\begin{equation}
  \mathrm{ED}(X,Y) \coloneqq \left|\frac{
  \frac{2}{nn'}\sum_{i,j}\lVert x_i-y_j\rVert
  -\frac{1}{n(n-1)}\sum_{i\neq i'}\lVert x_i-x_{i'}\rVert
  -\frac{1}{n'(n'-1)}\sum_{j\neq j'}\lVert y_j-y_{j'}\rVert}
  {\frac{1}{n'(n'-1)}\sum_{j\neq j'}\lVert y_j-y_{j'}\rVert}\right|^{1/2}.
\end{equation}
The numerator is the energy distance, which vanishes if and only if the two clouds are drawn from the same distribution; the denominator is the mean pairwise distance within the true
cloud, which makes $\mathrm{ED}$ dimensionless. ED is interpretable as the displacement between clouds as a fraction of the actual distributional width.
The self terms exclude the diagonal so that the numerator is unbiased, and is exactly
zero in expectation at any finite sample size when the predicted distribution is correct.
Reported on the 3D position of lost particles on the boundary surface
$x_i= (x(\theta, \varphi), y(\theta, \varphi), z(\theta, \varphi))$, ED measures deviation in the shape of the spatial distribution of lost particles. The metric is only computed on devices with at least 10 lost particles.

\item \textit{Heat Flux (HF).} A top-line physics metric measuring the accuracy of the
energy per unit area predictions on the boundary surface. We report the relative $L_2$ error in the heat flux,
\begin{equation}
  \mathrm{HF} \coloneqq \left(\frac{\int_S \left(Q_q-Q_p\right)^2 dA}{\int_S Q_{\text{ref}}^2\, dA}\right)^{1/2},
\end{equation}
where $p, q$ are the true and predicted distributions. The reference heat flux, $Q_{\text{ref}} = 3.52/\int_S dA \,\,\left[\frac{\unit{\mega\electronvolt}}{\unit{\meter}^2}\right]$, assumes all particles are lost with full energy, uniformly over the boundary. The integrals are computed by
binning particle loss position and energy samples onto an $m\times m$ grid over a single field period of $\theta,\varphi$ (an explicit formula is given in \Cref{app:heat_prediction}).
Leveraging the field periodicity of the equilibrium in the grid spacing gives $\nfp$ times more particles per bin. We report HF for $m=16,32$ to balance discretization and statistical error.

\item \textit{Loss Fraction (MAE, MedAE).} The median (MedAE) or mean (MAE) of the absolute error between the predicted loss fraction and the actual, taken over all devices and initial surfaces, $s_{\text{init}}$.
\end{enumerate}


\paragraph{Results}

\begin{table}[tbh!]
\centering
\caption{Evaluation of IID vs PointNet for stellarators from the \emph{test} set using particles born on $s=0.1,0.25,0.5$. HF and the loss-fraction errors are over the entire test set ($N=74763$); ED needs a lost-particle cloud and is restricted to devices with at least $10$ lost particles ($N=72512$). The headline results show that the IID model outperforms the PointNet model in terms of median energy distance and heat flux.  Lower is better for every reported metric. \textbf{Bold} = winner per row. The Monte Carlo floor (MC floor) denotes the median value of the metric under sampling noise variation when using the perfect predictor. Almost reaching the MC floor, the IID model demonstrates practical performance.}
\label{tab:iid_vs_pn}
\begin{tabular}{lccc}
\toprule
Metric & IID & PointNet & $\Delta$\% \\
\midrule
\multicolumn{4}{l}{\textit{Headline}} \\
ED (median) & \textbf{0.0306} & 0.0358 & +16.9 \\
\quad MC floor & 0.019 & 0.019 & - \\
HF $16\times16$ (median) & \textbf{0.3447} & 0.3824 & +10.9 \\
\quad MC floor & 0.187 & 0.187 & - \\
HF $32\times32$ (median) & \textbf{0.5720} & 0.6134 & +7.2 \\
\quad MC floor & 0.384 & 0.384 & - \\
Loss fraction Med-AE & 0.0186 & \textbf{0.0183} & -1.7 \\
\bottomrule
\end{tabular}
\end{table}

\begin{figure}
    \centering
    \includegraphics[width=\linewidth]{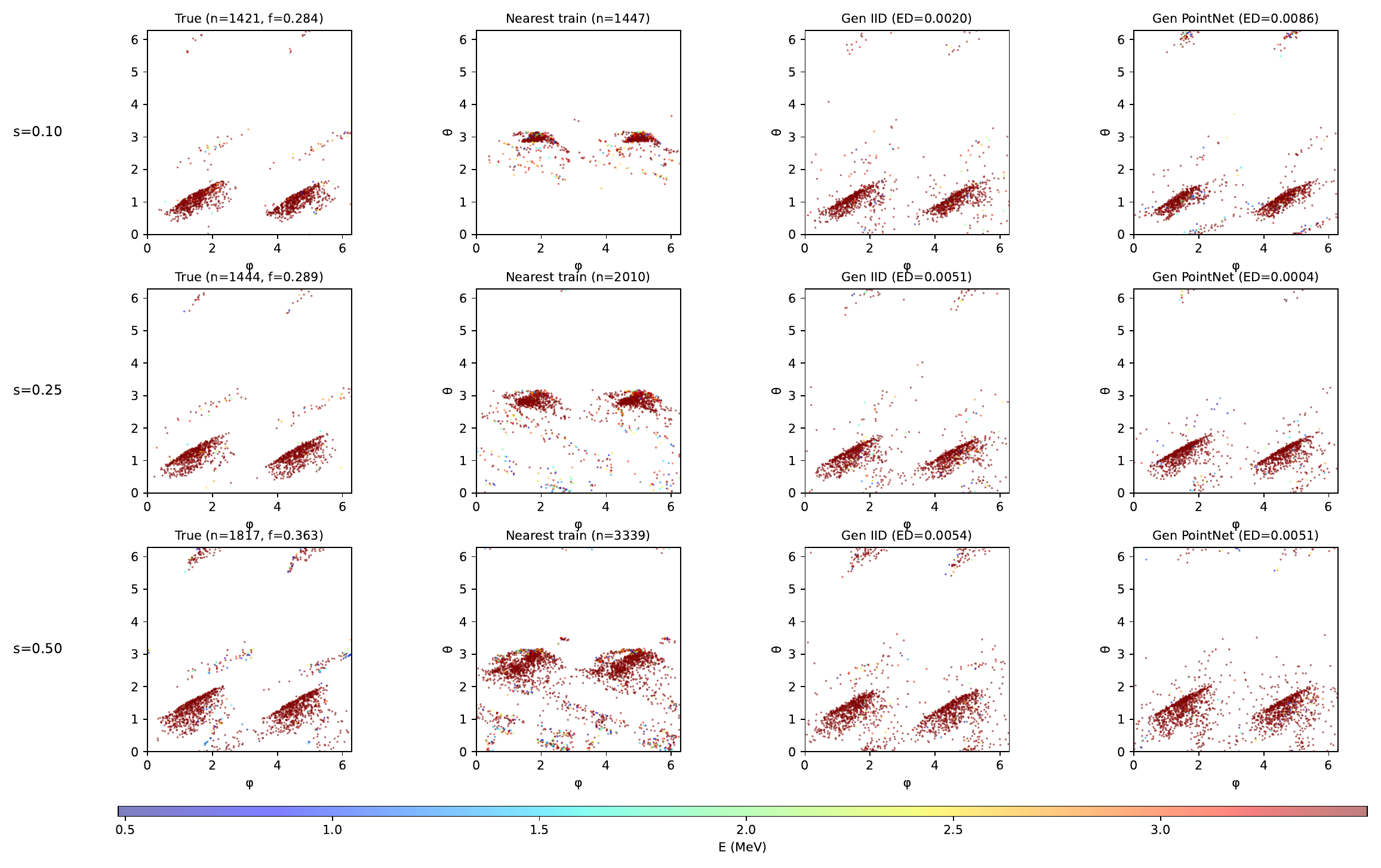}
     \includegraphics[width=\linewidth]{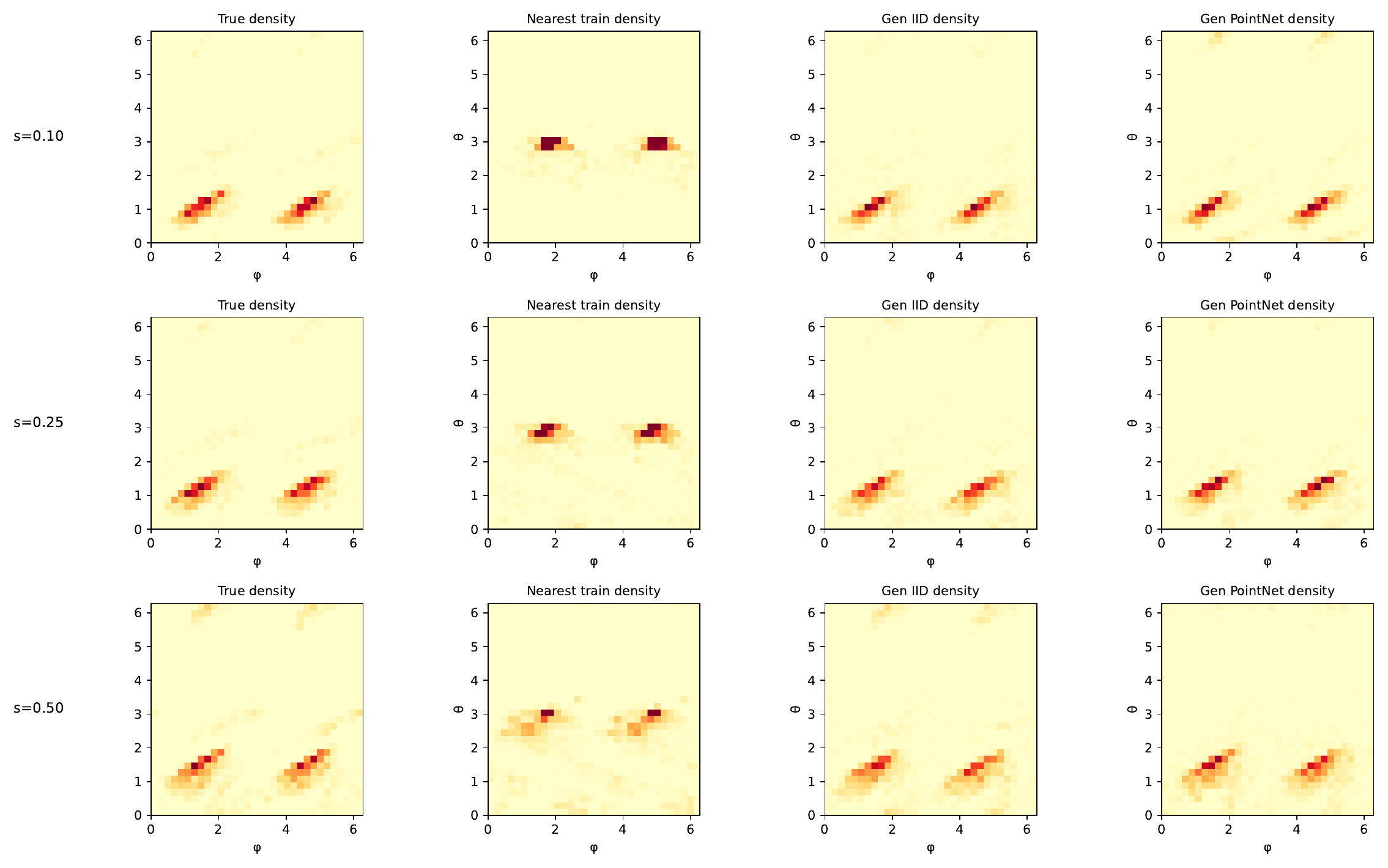}
    \caption{Heat deposition prediction in a single stellarator: (top 3 rows) lost particle point cloud position and energy (color) for born on surfaces $s=0.10, 0.25, 0.50$; (bottom 3 rows) lost particle density per surface $s=0.10, 0.25, 0.50$. Columns from left to right show the actual device, the nearest device to target from the training set (we compute the nearest device by first filtering devices with $[0.9\langle B\rangle, 1.1\langle B\rangle]$, followed by finding the nearest neighbor in the Euclidean norm of the flattened $\chib_s$ vector), the prediction from the IID baseline, and the prediction from the PointNet baseline.}
    \label{fig:heat_flux_examples}
\end{figure}

We report the quantitative evaluation metrics in \Cref{tab:iid_vs_pn}, and qualitative visualization on the predicted heat deposition pattern versus the truth of example devices from the validation set in \Cref{fig:heat_flux_examples}. We see that the heat deposition predictions from CFM show satisfactory performance, capturing the toroidal geometry from the ground truth while not memorizing from the training set. The IID baseline outperforms the PointNet baseline across most metrics (except loss-fraction Med-AE). While PointNet is a more flexible model, it is significantly more expensive to train and may perform worse at an equivalent training cost. Additional performance metrics can be found in \Cref{app:heat_prediction}. 

\subsubsection{Trajectory prediction}
\label{sec:traj_prediction}

Fast ions follow chaotic trajectories through the magnetic field, governed by the guiding center equations \Cref{eq:guiding_center}. While the loss fraction and heat deposition pattern summarize the outcome of a GC simulation, the trajectories themselves show \emph{how} and \emph{when} particles are lost: whether a particle drifts steadily outward or remains confined. A surrogate that rolls out trajectories conditioned on the device would serve as a fast stand-in for the GC integrator and can be formulated as an ML task as follows: given the device design $\chib$ and the initial guiding center state of a fast-ion $\zb_0$, predict the state $\zb(t)$ for all lead times $t \in [0, 1\unit{\milli\second}]$. Since the guiding center equations are deterministic, the target is a single trajectory; key challenges come from the sensitivity of trajectories to their initial conditions, phase transitions, high-frequency trajectories, long prediction horizons, and the need to generalize to unseen magnetic geometries. 

\paragraph{Model data}
We train on the roughly 22k devices in the \texttt{BoSS} with saved trajectories, each with 6000 particles born on the $s=0.1, 0.25, 0.5$ surfaces and sampled every $1\unit{\micro\second}$ until $1\unit{\milli\second}$ or until lost (\Cref{sec:pipeline}). Since every particle carries the same energy, we replace $\vpar$ with the pitch $\lambda = \vpar/v\in[-1,1]$, and embed the guiding center state $(s,\theta,\varphi,\lambda)$ as $\zb = (\sqrt{s}\cos\theta, \sqrt{s}\sin\theta, \cos\varphi, \sin\varphi, \lambda) \in \R^5$, which removes the angular branch cuts and the coordinate singularity at the magnetic axis. In Boozer coordinates the guiding center equations depend on the device primarily through the field strength $B(s,\theta,\varphi)$ and the profile functions (\Cref{app:guiding_center_equations}). We therefore set $\chib = (\bb, \iotab, \mathbf{G}, \Ib, \pb, \nfp, \psiedge) \in \R^{94}$, where $\bb\in\R^{32}$ holds the leading PCA coefficients of $B$ on a $15\times41\times41$ Boozer grid (PCA is performed on the collection of flattened vectors), $\Gb$ is a vector of the Boozer poloidal current profile $G(s)$ evaluations \citep{imbert-gerard_introduction_2024}, the profiles are evaluated on $15$ radial points, and $\psiedge$ is the edge toroidal flux. 

\paragraph{Model baselines}
We train three deterministic baselines of matched model size ($\approx 10$M parameters), described in \Cref{app:traj_prediction}: (a) AR-MLP, a memoryless one-step dense neural network $\hat{\zb}_{t+1} = f(\hat{\zb}_t, \chib)$ rolled out autoregressively; (b) AR-Transformer, a causal windowed transformer $\hat{\zb}_{t+1} = f(\hat{\zb}_{t-99:t}, \chib)$ that attends to the last $100$ states of its rollout and uses rotary position embeddings (RoPE) \citep{su_roformer_2024}; and (c) Direct-MLP, which predicts $\hat{\zb}(t) = F(\zb_0, t, \chib)$ in a single forward pass. The baselines are designed as controlled pairs: AR-MLP versus AR-Transformer isolates the value of memory, which is unnecessary in principle since $\zb$ is the complete state of an autonomous ODE but may compensate for the $1\unit{\micro\second}$ sampling; AR-MLP versus Direct-MLP isolates the effect of autoregression. All models predict $\varphi$ instead of $(\cos\varphi, \sin\varphi)$ and use an output head that keeps $(\cos\varphi, \sin\varphi)$ on the unit circle. Due to the computational burden of long rollouts, autoregressive models are trained on rollouts of up to $100\unit{\micro\second}$, while Direct-MLP is trained on lead times spanning the full $1\unit{\milli\second}$. The window length for the transformer was chosen to match the training horizon so that the model input is never an unseen length. The autoregressive models are trained on sections of successive points from the trajectory, where the initial time is chosen uniformly over $1$ to $999$. We also report two reference predictors: \textit{stationary}, which predicts $\hat \zb(t) = \zb_0$, and \textit{device mean}, which predicts the mean state of each device's particles over all time.

\paragraph{Evaluation metrics}
Models are evaluated on devices from the test split. 
Due to the high cost of model evaluation, evaluation is performed on $n_p=500$ randomly selected trajectories from each of the $n_d=3184$ test devices or an equivalent number of trajectories and devices from the held-out training data; these values ensure that our metrics are statistically valid. A predicted particle is lost the first time its predicted $s$ crosses $s=0.98$. The loss times for the autoregressive models are rounded to the nearest microsecond increment, while the loss time is computed via bisection search for Direct-MLP. During evaluation, model predictions are held fixed for all time after the particle is lost. 

\begin{enumerate}
\item \textit{Trajectories (nRMSE).} The prediction error at lead time $t$, for each channel $g$ of $\zb$,
\begin{equation}
\mathrm{nRMSE}_g(t) \coloneqq \frac{1}{\sigma_g}\,\frac{1}{n_d}\sum_{d=1}^{n_d}\left(\frac{1}{n_p}\sum_{i=1}^{n_p}\big\lVert  
\hat z_{d,i,g}(t) - z_{d,i,g}(t)\big\rVert^2\right)^{1/2},
\end{equation}
where $z_{d,i,g}$ is channel $g$ of particle $i$ on device $d$, $\hat{z}_{d,i,g}$ is the predicted value and $\sigma_g^2 = \mathbb{E}\,\lVert z_g\rVert^2$ is the second moment of channel $g$ over the
training set (pooled over devices, particles, and lead times). The average is taken over all particles that are not lost by time $t$. 
The channels considered are $\sqrt{s}\cos(\theta)$,$\sqrt{s}\sin(\theta)$, $\lambda$, and the pair $(\cos\varphi,\sin\varphi)$.
Unlike the other channels, which can be interpreted as a percent error, the toroidal angle channel has an nRMSE of $\sqrt{2}$ when the prediction is uncorrelated -- this follows
by expressing the squared error as $2\big(1-\cos(\hat\varphi-\varphi)\big)$, and noting that $\sigma_g=1$ when $\varphi, \hat\varphi$ are uncorrelated. The total nRMSE is the average over the channel-level nRMSE.

\item \textit{Loss fraction (MAE).} The mean over devices of the absolute prediction error in the loss fraction. 

\item \textit{Confinement time ($W_1$).} The Wasserstein-1 distance (measured in microseconds) between the predicted and true confinement time distributions $p(T | \chib)$. The reported metric averages the per-device distance over all test devices.
\end{enumerate}

\begin{figure}
    \centering
    \includegraphics[width=\linewidth]{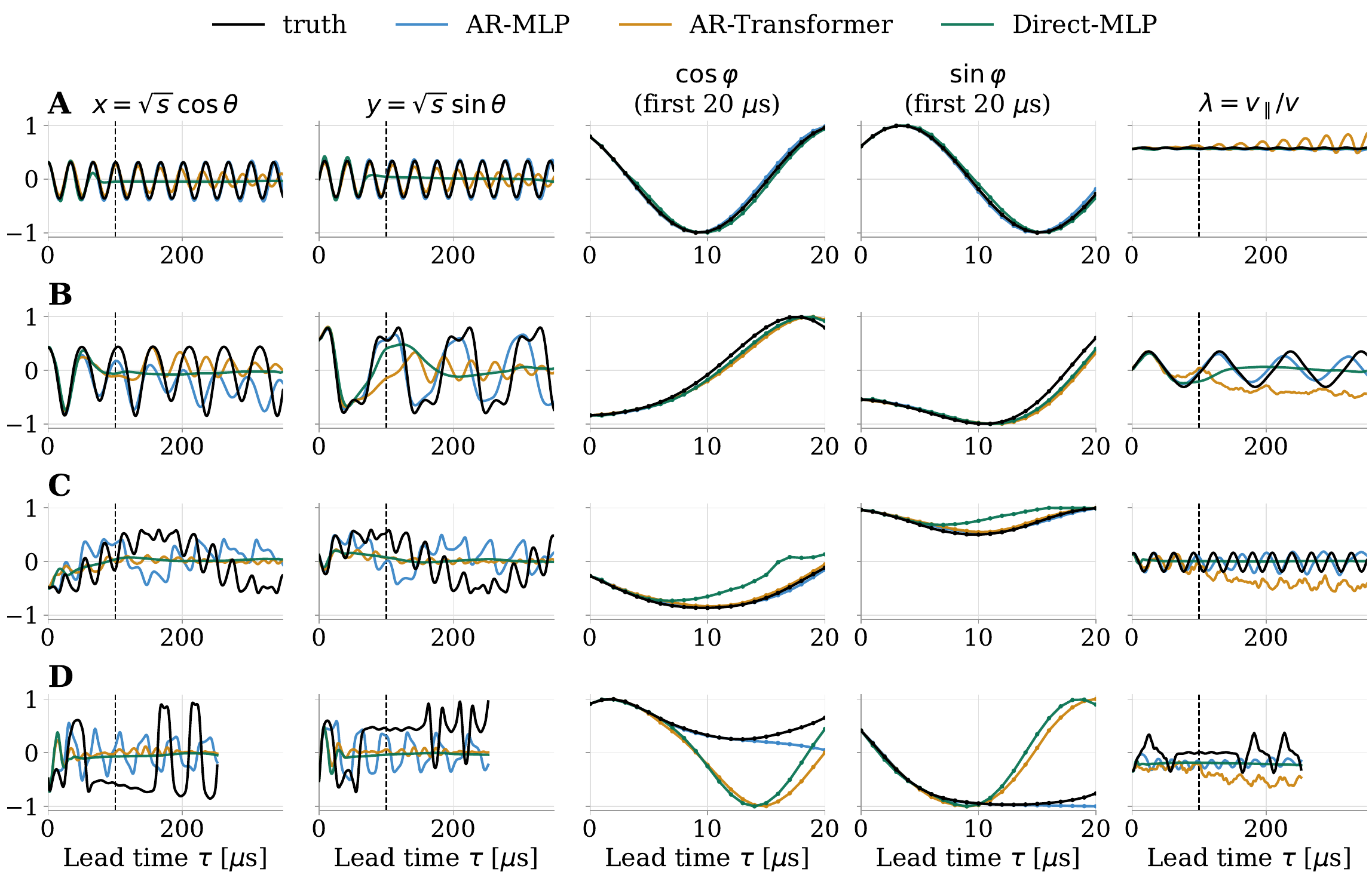}
    \caption{Example predictions of the three trajectory models over time for distinct trajectories (rows) across the five channels (columns). Trajectory A is a regular single frequency trajectory over the horizon, while B and C display multiple frequencies, and D undergoes a phase transition. The vertical dashed line indicates the training horizon for the autoregressive models (AR-MLP, AR-Transformer).}
    \label{fig:traj_pred_rollouts}
\end{figure}

\paragraph{Results}
\Cref{fig:traj_pred_rollouts} shows example rollouts, and \Cref{tab:traj_pred_results} summarizes the performance metrics on the \emph{test} set. A plot separating the nRMSE versus lead time by channel can be found in \Cref{app:traj_prediction}. All three models are far more accurate than the reference predictors at short lead times, but their errors grow steadily with lead time, and no model predicts any channel reliably to $1\unit{\milli\second}$. \Cref{tab:traj_pred_results} shows that AR-MLP often predicts loss fraction within ten percentage points, despite having poor accuracy on long-horizon trajectory prediction, indicating that a model with inaccurate trajectory predictions may still be able to predict statistics.

Within the $10\unit{\micro\second}$ window, AR-MLP outperforms the transformer in terms of nRMSE, but is outperformed on the mid- and long-horizon predictions. Direct-MLP is the least accurate model at short lead times but is the only model whose error stays below that of the device mean at long lead times, since its errors do not compound. Long horizon prediction of both autoregressive models could be improved by extending the training window to capture the full $1\unit{\milli\second}$ trajectories, though at a significant computational cost. AR-Transformer training cost was dominated by backpropagating through the $K$-step rollout, which made long rollouts increasingly expensive. Despite using roughly double the wallclock time than AR-MLP, AR-Transformer saw significantly less data during training. 

In \Cref{app:traj_prediction}, we perform an ablation study on the training horizon and find that increasing the training horizon extends the validity of model predictions to the longer horizon without sacrificing performance on the shorter horizons. This suggests that extending the training horizon to penalize rollout error on full-length trajectories could significantly improve performance. However, doing so requires rethinking model architecture and the training procedure in order to reduce training costs.

This benchmark emphasizes a need for computationally efficient methods for long-horizon problems and models for trajectories with phase transitions and rapid oscillations, as shown in \Cref{fig:traj_pred_rollouts}C. The periodic structure of the data visible in \Cref{fig:traj_pred_rollouts} emphasizes an opportunity to evaluate specialized baselines such as learning effective dynamics \citep{vlachas2022multiscale} or develop structure-exploiting models.

\begin{table}[t]
\centering
\caption{Trajectory prediction metrics on the \emph{test} set. Stationary predicts the initial condition, $\hat \zb(t) = \zb_0$, and so never loses a particle. Standard errors of each metric, computed via bootstrap subsampling, are below $0.0004$ for nRMSE and loss fraction MAE and below $0.3\,\unit{\micro\second}$ for $W_1$. Lower is better for every metric. \textbf{Bold} = best per row.}
\label{tab:traj_pred_results}
\begin{tabular}{lcccc}
\toprule
Metric & Stationary & AR-MLP & AR-Transformer & Direct-MLP \\
\midrule
nRMSE ($\tau = 10\,\mu$s) & 1.019 & \textbf{0.261} & 0.378 & 0.416 \\
nRMSE ($\tau = 100\,\mu$s) & 1.042 & 0.846 & 0.807 & \textbf{0.761} \\
nRMSE ($\tau = 999\,\mu$s) & 1.038 & 1.046 & 1.094 & \textbf{0.862} \\
Loss fraction MAE & 0.183 & \textbf{0.104} & 0.118 & 0.137 \\
Confinement time $W_1$ [$\mu$s] & 159.8 & \textbf{79.4} & 105.2 & 115.5 \\
\bottomrule
\end{tabular}
\end{table}

\section{Conclusion and Future Work}
\label{sec:conclusion}
We introduce the \texttt{BoSS}, a large-scale dataset of diverse stellarator designs with high-fidelity fast-ion simulations, and baselines for three practical tasks with evaluation protocols.
The \texttt{BoSS} provides new avenues for the ML community to accelerate stellarator design. In particular, ML surrogates for fast ion loss statistics enable rapid evaluation of stellarator designs, and generative models for stellarator design facilitate a fast exploration of the design space, revealing novel and practical designs.

The \texttt{BoSS} leaves room for improvement in at least four directions, including data, task, model, and evaluation. On the data side, incorporating a more comprehensive range of physics simulations would increase the applicability of ML methods for stellarator design. Future datasets could include magnetic fields produced by both coils and plasma currents, fast-ion simulations with Coulomb collisions, or data from gyrokinetic and neutronics simulations. Moreover, the process of data collection could be streamlined by ``closing" the active learning loop: surrogates could be trained on the active learning data and then used once again to collect more data. Surrogates with calibrated uncertainty estimates, such as Gaussian processes, are attractive candidates since they simplify the process of estimating the prediction error. On the task side, future work could go beyond magnetic field design to address stellarator engineering and construction, for instance by including coil sets and vacuum vessel geometries in a dataset. 
On the modeling side, black-box ML models could be replaced or enhanced with physics-informed methods, which have the potential to substantially push benchmark performance and improve interpretability. Opportunities for physics-informed ML are available in trajectory prediction, where conserved quantities and trajectory structure are known, and in inverse design, where engineering constraints such as the mirror ratio could be enforced exactly rather than through penalties or conditioning. 

Finally, automated evaluation pipelines could be developed for a range of tasks to ease usage and ensure verifiability of results. Connecting two fields such as ML and plasma physics is difficult because of a communication gap between the fields. A benchmark helps bridge that
gap by not only providing cross-discipline tasks but also providing a means of checking claims about model predictions or results against the underlying physics. Our evaluation pipeline and the task-specific metrics are one such check. 
Future work could take this a step further: packaging the stellarator design pipeline and metrics as an openly available service would let ML researchers validate their designs without running physics codes, improving
accessibility for the ML community while ensuring that claims of physical significance are verifiable and not overstated. In addition, an automated pipeline would could integrate seemlessly with optimization algorithms allowing for automated, closed-loop stellarator design. Complementarily, better integration between ML models and established plasma physics codes would allow users to quickly extract data-driven insights.

\impact{
By publishing scientific data and evaluation protocols, our benchmark creates an opportunity for ML methods to impact science in a positive way, through accelerating fusion research. Conversely, the benchmark may also have a potential positive impact on ML by grounding method development in practical scientific problems. Potential negative impacts stem from overstatement and misinterpretation of results. In particular, the tasks in our benchmark focus on design of a stellarator magnetic field with the aim of improving one dimension of device performance, guiding center confinement. While progress on these tasks represents a meaningful step toward advancing fusion research, it does not address important challenges in coil design, engineering, or other dimensions of the physics.
Moreover, solving these tasks with ML surrogate or generative models does not necessarily imply that the learned models capture the underlying physical mechanisms, nor that the models can generalize out-of-sample to arbitrary stellarator designs. To mitigate these risks, we emphasize clear scoping and careful exposition. We encourage future work to expand the benchmark to capture a wholistic view of stellarator design, and develop physics-informed ML models for fusion.
}

\section{Code availability}
Code for generating the data and results will be available upon publication. The data is available through \href{https://boss.flatironinstitute.org/}{https://boss.flatironinstitute.org}.

\section{Disclosure on the use of artificial intelligence}
The authors acknowledge the use of large language models for developing scripts, running experiments, and reviewing and editing the manuscript.

\acks{The authors would like to thank the Flatiron Institute’s Scientific Computing Core, and the Simons Foundation for their support. We appreciate Rogerio Jorge, Andrew Giuliani, and for discussions and feedback, Tobias Blickhan, Vlado Menkovski, Madhi Mehmanchi, and Federico Danieli for insights into trajectory modeling, Rahul Gaur for sharing data from the Omnigenity database. 
We thank Qin Li and the University of Wisconsin-Madison SciML for Plasma workshop for providing a venue for us to communicate this work. We thank Lucy Reading-Ikkanda for the logo design.}

\bibliography{references_cleaned}

\clearpage
\appendix
\crefalias{section}{appendix}

\section[Appendix]{Surface description}
\label{app:surfaces}
The plasma boundary is parameterized as a Fourier series in the \texttt{SurfaceRZFourier} format, the standard description used by \texttt{VMEC} and \texttt{DESC}. The cylindrical coordinates $R,Z$ are expressed using an arbitrary poloidal angle $\theta$ and standard cylindrical angles $\phi$ as,
\begin{equation}
\begin{split} 
    R(\theta,\phi) &= \sum_{n=0}^{\ntor} R_{0,n}\cos(- n_{\text{fp}}n\phi) + \sum_{m =1}^{\mpol} \sum_{n=-\ntor}^{\ntor} R_{m,n}\cos(m\theta - n_{\text{fp}}n\phi),
    \\
    Z(\theta,\phi) &= \sum_{n=1}^{\ntor} Z_{0,n}\sin(- n_{\text{fp}}n\phi)  + \sum_{m =1}^{\mpol} \sum_{n=-\ntor}^{\ntor} Z_{m,n}\sin(m\theta - n_{\text{fp}}n\phi),
\end{split}
    \label{eq:vmec_fourier_rep}
\end{equation}
where $R_{m,n}, Z_{m,n}$ are Fourier coefficients representing the surface, and $\ntor,\mpol = \{0,1,2,\ldots\}$ can be increased to achieve more complicated boundary representations. Field period symmetry with $\nfp$ periods and stellarator symmetry have been assumed in this representation. The values of $\mpol, \ntor$ vary across the dataset. $R_{00}$ is the major radius.

\subsection[Appendix]{Randomly sampling stellarator boundary shapes}
\label{sec:random_sampler}
Random boundaries are generated by fitting a truncated Fourier surface to points sampled at a random radius about a random magnetic axis. The construction takes six parameters: the number of field periods $n_\text{fp}$, Fourier resolution $M$,
target minor radius $a$, aspect ratio $A$ (so that $R_0 = aA$), axis amplitude $\alpha$ controlling the excursion of the magnetic axis, and a shaping amplitude $\sigma \ge 1$ controlling how non-circular the cross sections are. Setting $\alpha = 0$ and $\sigma = 1$ removes both sources of randomness and recovers an axisymmetric circular torus of minor radius $a$ and major radius $R_0$.

Quadrature points on the boundary are generated on a uniform grid of $N = 4M + 2$ points per direction,
$\theta_k = 2\pi k/N$ and $\phi_j = 2\pi j/(N n_\text{fp})$ for $j,k = 0,\dots,N-1$, covering
one field period in $\phi$. At each toroidal grid point the magnetic axis is drawn
independently as
\begin{equation}
  R_\text{ax}(\phi_j) = R_0 + \tfrac{\alpha a}{2} u_j,
  \qquad
  Z_\text{ax}(\phi_j) = \tfrac{\alpha a}{2} v_j,
  \qquad u_j, v_j \sim \mathcal{U}[-1,1],
\end{equation}
and a boundary point is placed at an independently drawn radius
$r_{jk} \sim \mathcal{U}[a/\sigma,\, \sigma a]$ about that axis,
\begin{equation}
  R_{jk} = R_\text{ax}(\phi_j) + r_{jk} \cos \theta_k,
  \qquad
  Z_{jk} = Z_\text{ax}(\phi_j) + r_{jk} \sin \theta_k.
\end{equation}

A stellarator-symmetric boundary of the form \Cref{eq:vmec_fourier_rep} using $\mpol=\ntor = M$ is then fit to $\{(R_{jk}, Z_{jk})\}$ in the least-squares sense. Because the grid is uniform and $N > 2M+1$, the retained harmonics are discretely orthogonal, so the fit is exactly the truncated discrete Fourier projection of the sampled data. Finally the boundary is shifted such that $R_{00}=R_0$ and every remaining coefficient is scaled to achieve the desired minor radius, $a$. 

\subsection[Appendix]{Measuring the distance between surfaces}
\label{app:distance_metric}

\begin{figure}[tb!]
    \centering
    \makebox[\textwidth][c]{%
    \includegraphics[width=0.9\textwidth]{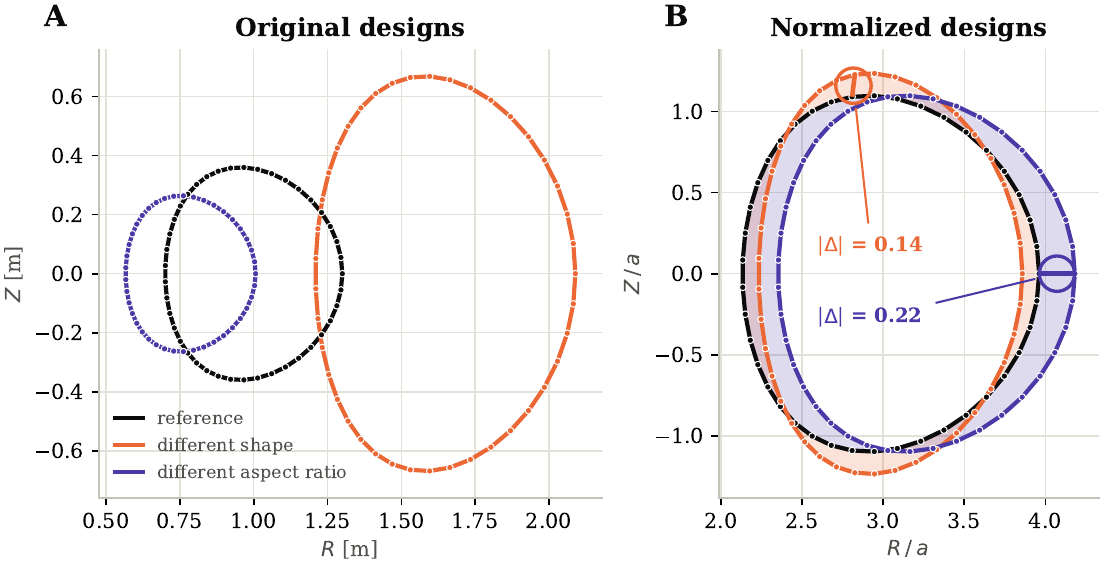}
    }
    \caption{Graphic illustrating the normalized distance. Frame A shows the $\phi=0$ cross section of three surfaces: a reference surface (black), a surface with distinct shape (orange), and a surface with distinct aspect ratio (violet). Frame B shows the normalized cross sections between which the distance is calculated.}
    \label{fig:aid_explainer}
\end{figure}

When using ML models to generate new surface designs it is useful to quantify design novelty in terms of the distance between the generated surface and the surfaces in the training set. We define a scale-free distance that accounts for the arbitary choice of $\theta$ (location of $\theta=0$, rotation direction, and speed). 
To compute the distance, the surfaces are first canonicalized to have a standard poloidal angle, and rescaled to have unit minor radius. Then the cylindrical components $R_{ij} = R(\phi_i, \theta_j),Z_{ij}=Z(\phi_i, \theta_j)$ are computed on a uniform grid in $\phi,\theta$. Finally, the Euclidean distance is computed between corresponding quadrature points, and the infinity norm is reported,
\begin{align}
    d(S, S') = \max_{ij}\sqrt{\left(\frac{R_{ij}}{a}  - \frac{R_{ij}'}{a'}\right)^2 + \left(\frac{Z_{ij}}{a} - \frac{Z_{ij}'}{a'}\right)^2},
\end{align}
where $a,R_{00}$ denote the minor and major radii. When comparing two stellarator symmetric surfaces, the field period can begin at $\phi=0,2\pi/\nfp$; in this case, we compute the distances between the surfaces and their mirrors at the two angles and report the minimum.  

A graphic illustrating the distance is shown in \Cref{fig:aid_explainer}. Frame A shows three surfaces which, in Frame B, are normalized to unit minor radius prior to undergoing the distance computation. Because points are paired by arc length rather than by nearest neighbor, the displacement need not be normal to either boundary. 

\paragraph{Parameterization of $\theta$}
The \texttt{SurfaceRZFourier} format allows for any poloidal angle $\theta$ to be used. For distance computations we choose $\theta$ to have the unit-speed parameterization, with $\theta =0$ occuring at the outboard midplane point: the $\theta$ maximizing $R(\theta)$ subject to $Z(\theta) = \langle Z\rangle$. The optimization problem can be solved globally by rewriting the trigonometric polynomials as algebraic polynomials -- using the substitution $w = e^{i\theta}$ -- and finding the exact roots of the polynomial through the companion matrix method. The traversal direction is selected so the sign of $\frac{dZ}{d\theta}(0)$ is positive.

\section[Appendix]{Guiding center equations in Boozer coordinates}
\label{app:guiding_center_equations}
The guiding center equation \Cref{eq:guiding_center} can be expressed in Boozer coordinates $(s,\theta, \varphi)$ as
\begin{align}
\begin{split}
    \dot s &= m\frac{\vpar^2/B + \mu}{\iota D \psi_0}\left(I B_\varphi - G B_\theta\right)
    \\
    \dot \theta &= \left( \left(G B_\psi - K B_\varphi \right) m(v_{||}^2/B + \mu) - C \vpar B \right)/(\iota D)
    \\
    \dot \varphi &= \left(F \vpar B - \left(B_\psi I - B_\theta K \right) m(\vpar^2 /B + \mu) \right)/(\iota D)
    \\
    \dot v_{||} &= (CB_\theta - FB_\varphi)\mu B/(\iota D)
    \\
    C &= - m v_{||} K_\varphi/B  - q \iota + m \vpar G_\psi/|B|
    \\
    F &= - m v_{||} K_\theta/B + q + m \vpar I_\psi /|B|
    \\
    D &= (F G - C I))/\iota
\end{split}
\label{eq:gc_motion_cart}
\end{align}
Here, $t$ is time, $m$ is the particle's mass, $q$ is the particle's charge, $B=|\Bb|$ is the field strength, $\mathbf{b}=\Bb/B$, $2\pi\psi_0$ is the toroidal flux at the plasma boundary, $G(s), I(s)$ are 
the poloidal and toroidal current times $\mu_0/(2\pi)$, $K$ is Boozer's $K$, $\iota(s)$ is the rotational transform, $\vpar$ and $\vperp$ are the components of velocity parallel and perpendicular to $\Bb$. Subscripts $\psi, \theta, \varphi$ are used to denote derivatives, where $s = \psi/\psi_0$. See \citep{imbert-gerard_introduction_2024} for a review of Boozer coordinates and \citep{de_blank_guiding_2000} for a discussion of the guiding center equations.

\section[Appendix]{Expanding the dataset with active learning}
\label{app:active_learning}

This appendix provides further details on the rejection sampling and optimization methods used to collect new data points as part of the active learning procedure.

In the random sampling approach, stellarator geometries $\chib_s$ are randomly sampled for each $\nfp=1,\ldots,6$, using the algorithm from \Cref{app:surfaces}, $e_A(\chib_s)$ is evaluated, and \texttt{VMEC} is evaluated assuming vacuum profiles, $\pb = \Ib = 0$. If $e_A(\chib_s) \geq \tau$ and \texttt{VMEC} converges, the point is saved. $\tau=0.5\%,1\%$ were used in practice, as depicted in \Cref{fig:active_learning}A. The random devices have minor radius uniformly sampled from $[0.5,2.0]$ meters, and aspect ratio uniformly sampled from $[3.5, 9.5]$.

In the optimization approach, data points are selected by minimizing $J(\chib)- \gamma e_A(\chib)^2$
with L-BFGS-B through scipy \citep{zhu_algorithm_1997,virtanen_scipy_2020}, where $J$ is described in \Cref{eq:opt_problem}. The objective is similar to the lower confidence bound acquisition function from the Bayesian optimization literature \cite{srinivas_gaussian_2009,frazier_tutorial_2018}, and seeks points with low $J$ or potentially inaccurate predictions.
For these optimizations, the pressure profile was set to $p(s) = \frac{3}{5}\frac{\langle B\rangle^2  \beta}{\mu_0} (2 - 3s + s^2)$ where $\langle B\rangle$ is drawn uniformly from $[4,10]$, $\beta\in\{0.0, 0.01, 0.02, 0.03, 0.04\}$. The target aspect ratio and minor radius for the optimization were randomly selected, and the mirror ratio target was randomly selected uniformly in $[1.2, 2.0]$. Initial candidates, for the optimization are either randomly sampled using the procedure from \Cref{app:surfaces}, or taken from the \texttt{BoSS}. Stellarators produced by these two methods are passed through the pipeline in \Cref{fig:pipeline}, and added to the dataset.

\paragraph{Elongation} 
\Cref{fig:active_learning}C plots the mean elongation of devices. The elongation of a constant $\phi$ cross-section of a surface is defined as the ratio of the extents along the two principal directions. This equals 1 for a circle and $b/a$ for an ellipse with semi-axes $b \ge a$, at any orientation. The elongation is a key dimension of variation for toroidal surface geometries \citep{landreman_direct_2018,giuliani_comprehensive_2025,rodriguez_near-axis_2025}. We report the mean elongation over all cross sections in a field period.

To compute the principal directions of a cross section, the cross section is discretized at 256 points uniformly in arclength, the $(R,Z)$ coordinates are translated so that the area centroid of the cross section lies at the origin, and the singular value decomposition of the second moment matrix (centered about the centroid, not the mean) is computed. The singular vectors $\vb_1(\phi),\vb_2(\phi)$ of the covariance are the principal directions, with the larger singular value belonging to $\vb_1$. Each point on the cross section $(R(\theta,\phi), Z(\theta,\phi))$ can be projected onto the principal directions to produce coordinates $(\alpha_1(\theta,\phi),\alpha_2(\theta,\phi))$.
The elongation of the cross section is the ratio of its extent along the major axis to its extent along the minor axis,
\begin{equation}
    \kappa(\phi) \coloneqq \frac{\max_\theta \alpha_1(\theta,\phi) - \min_\theta \alpha_1(\theta,\phi)}{\max_\theta \alpha_2(\theta,\phi) - \min_\theta \alpha_2(\theta,\phi)}.
\end{equation}


\section[Appendix]{Inverse generation with diffusion models}
\label{app:diffusion}
This appendix provides additional details on the training and evaluation of the Diffusion baseline for the inverse generation task. 

\paragraph{Task Specification.} We state inverse design as a
generative modeling problem as follows:
given a vector of prescribed properties, sample a plasma
boundary whose MHD equilibrium approximately realizes those properties. 
The outputs of the generative model can be scored by re-running the physics (\texttt{VMEC} and guiding
center tracing), evaluating whether the properties of the surface
truly match those specified by the conditions. For non-physical or highly shaped input surfaces, \texttt{VMEC} may fail, in which case we report failures.

\paragraph{Representing the surface.}
A boundary is stored as the Cartesian positions of a $41 \times 41$ grid of points covering one field period of the surface. We convert it to cylindrical coordinates
$\chib_{ij} = \big(R(\phi_i, \theta_j),\, Z(\phi_i, \theta_j)\big)$,
with the cylindrical angle $\phi_i = 2\pi i/(41\,\nfp)$, $i = 0,\dots,40$,
spanning one field period $[0, 2\pi/\nfp)$ with the endpoint excluded, and the
poloidal angle $\theta_j = 2\pi j/41$, $j = 0,\dots,40$, spanning $[0,
2\pi)$ with the endpoint excluded (the \texttt{SIMSOPT} quadrature convention).
Thus, the array is a two-channel image on a torus with $41 \cdot 41 \cdot 2 =
3362$ entries.
\footnote{For reference, the common academic-sized natural-image generative modeling benchmark contains images of $32 \cdot 32 \cdot 3 = 3072$ dimensions.}
Surface coordinates are standardized with per-coordinate $z$-scoring.

\paragraph{Conditions.}
The model is conditioned on
\[
  \yb = \big(\nfp,\; \lossfrac,\; \psiedge,\; \aminor,\; A,\; \Bmin,\; \rmirror,\;
  \pb,\; \iotab,\; n_{\text{curr}}\big) \in \mathbb{R}^{38},
\]
whose entries are, in order: the number of field periods $\nfp \in
\{1,\dots,6\}$; the fast-ion loss fraction $\lossfrac$ for particles born on the $s = 0.1$ surface (the quantity defined in the dataset section); the
toroidal flux through the boundary $\psiedge$ (the \texttt{VMEC} input
\texttt{phiedge}, in Wb); the minor radius $\aminor$ (m); the aspect ratio $A$; the minimum of $|\Bb|$, $\Bmin$ (T); the mirror ratio $\rmirror
= B_{\max}/\Bmin$; the pressure profile $\pb \in \mathbb{R}^{15}$ and
rotational-transform profile $\iotab \in \mathbb{R}^{15}$, each evaluated on the same 15 uniformly spaced points in normalized toroidal flux $s \in
[10^{-4}, 1]$ used by the dataset; and the \texttt{VMEC} constraint flag
$n_{\text{curr}} \in \{0, 1\}$, where $n_{\text{curr}} = 0$ means the
equilibrium was computed with the rotational transform profile prescribed and $n_{\text{curr}} = 1$ means it was computed with the toroidal current profile prescribed.
When paired with the generated surface, $(\pb, \iotab, n_{\text{curr}}, \psiedge, \nfp)$ provides sufficient information to solve an equilibrium and run a guiding center simulation. We condition on $\psiedge$ rather than on $\langle B \rangle$ because $\psiedge$ is an input of the equilibrium solve.


\paragraph{Preprocessing of the conditioning vector.}
$\nfp$ is passed as an integer and mapped to a learned 8-dimensional embedding inside the network, since it is a symmetry label rather than a magnitude. The
six scalars $\lossfrac, \psiedge, \aminor, A, \Bmin, \rmirror$ are standardized via $z$-scoring with training-set means and standard deviations. The two profiles are transformed by $x \mapsto \operatorname{arcsinh}(x / c)$, with $c$ the
training-set standard deviation of the block, and then $z$-scored per $s$-point; the transform is linear near zero and logarithmic for large values, which tames
the heavy tail of the pressure profile (its values span several orders of magnitude) without amplifying near-zero noise. Pressure statistics are computed
over the non-vacuum devices only (roughly 35\% of the data), so that the 65\% of rows with $\pb \equiv 0$ map to a fixed constant rather than dragging the mean
to zero; rotational-transform statistics are computed over all rows. The flag $n_{\text{curr}}$ is passed as a raw $0/1$ value.

\paragraph{Flow Matching.}
The Diffusion model is trained via flow matching \citep{lipman_flow_2023,
albergo_stochastic_2023, liu_flow_2023}. 
Flow matching defines a time-dependent velocity field $\vb(t, \yb, \cdot )$ depending on conditions $\yb$.  
The velocity field is constructed such that the ODE $d \mathbf{X}_t = \vb dt$, with initial condition $\mathbf{X}_0 = \epsb\sim \mathcal{N}(0,
\mathrm{Id})$, has the time-$1$ solution $\mathbf{X}_1\sim p(\chib | \yb)$.
A velocity model $\vb_\omega$ parameterized by a neural network is trained to match the idealized velocity.
Under linear flow paths, $\xb_t = t\,\chib +  (1-t)\, \epsb$, one can show it is sufficient to minimize the mean squared error $\mathbb{E}\,\|\vb_\omega(t, \yb, \xb_t) - (\chib - \epsb)\|^2$. The neural network for the velocity -- acting on noisy $(R,Z)$ images, conditioning values $\yb$, and diffusion time $t$ -- is a 2D UNet. 

\paragraph{UNet Details.}
The network is a two-dimensional UNet
acting on the $(R, Z)$ image $\chib$. The $41 \times 41$ grid is reflect-padded to $44
\times 44$ so that it downsamples cleanly, and the network runs at resolutions
$44 \to 22 \to 11$ with 128 base channels, channel multipliers $(1, 2, 2)$, four
residual blocks per level in both the down and up paths, and multi-head
self-attention (4 heads) at the coarsest level and in the middle block;
downsampling is by strided convolution and upsampling by nearest-neighbour
interpolation followed by a convolution. The output is cropped back to $41
\times 41$. Conditioning enters through FiLM \citep{perez_film_2018}: a
128-dimensional sinusoidal embedding of $t$, the $\nfp$ embedding, and the 37
remaining entries of $\yb$ (the standardized scalars and profiles, and the raw
$n_{\text{curr}}$ flag) are concatenated and passed through a
two-layer MLP to a 512-dimensional conditioning vector, from which every
residual block predicts a per-channel scale and shift, applied as $(1 +
\text{scale}) \cdot \mathrm{GN}(h) + \text{shift}$ to the output of its second
group normalization. The FiLM projections are zero-initialized, so every block
starts out unconditioned, and the output convolution is zero-initialized, so the
initial velocity prediction is zero; both pathways are learned from scratch. The
model has 35.6 million parameters.

\paragraph{Training.} The model is trained on all 154{,}128 training devices with AdamW (no weight
decay), a learning rate of $2 \times 10^{-4}$ held constant after a 100-epoch linear warm-up, and a batch of 1024 (256 per GPU on four A100 GPUs, data-parallel) for 5{,}000 epochs.  An exponential moving average (EMA) of the weights with decay 0.999 is maintained and used for all sampling. All results in this section use the EMA weights at epoch 4000.

\paragraph{Sampling and post-processing.}
A design is generated by drawing $\Xb \sim \mathcal{N}(0, \mathrm{Id})$ and integrating $\dot{\Xb} = \vb_\omega(t, \yb, \Xb)$ from $t = 10^{-4}$ to $t = 1 -10^{-4}$ with a second-order midpoint scheme in 200 steps, then undoing the
per-coordinate standardization to obtain the $(R, Z)$ grid. The raw grid carries pointwise jitter at the resolution of the grid. The raw surface coordinates are projected by least squares onto a
stellarator-symmetric \texttt{SurfaceRZFourier} with poloidal and toroidal mode numbers $m \le 10$, $|n| \le 10$; this step smooths the surface, imposes stellarator symmetry, and produces the Fourier representation \texttt{VMEC} takes as its boundary.
The raw sampled grid is kept alongside the fitted surface,
so geometric fidelity (for instance, how closely the minor radius of the raw output matches $\aminor$) can be measured before smoothing.

\paragraph{Evaluation Protocol.} 
To evaluate a design, we take the conditioning vector $\yb$ used to generate the design, and re-run the forward pipeline that produced the dataset with $\chib_s, \yb$ as inputs. The fitted boundary is passed to fixed-boundary \texttt{VMEC} with the prescribed $\psiedge$, the prescribed $\pb$ (cubic spline in $s$), and, according to $n_{\text{curr}}$, either the prescribed $\iotab$ or the (zero) current profile. \texttt{VMEC} is run with the dataset's
most common resolution parameters ($\texttt{MPOL,NTOR}=10$, radial multigrid $\texttt{NS}=16/32/64/101$ surfaces, $\texttt{FTOL}=10^{-13}$). From the converged equilibrium, we trace fast ions exactly as in the dataset generation: a Boozer-coordinate interpolant of the field (32 poloidal and toroidal modes, $25^3$ grid), 15{,}000 alpha particles born uniformly on the $s= 0.1, 0.25, 0.5$ surfaces with isotropic velocities, traced with \texttt{CATAPULT} through \texttt{FIRM3D} to $\tmax = 0.1\unit{\sec}$.
The reported loss fraction is the fraction of the 5{,}000 particles born on $s = 0.1$ that are lost before $\tmax$. This protocol also
exposes a failure mode with no analogue in image generation: on some generated boundaries the equilibrium solver does not converge. 

\subsection[Appendix]{Results at the ARIES-CS scale}
Fast-ion confinement depends on device size and field strength as well as on shaping. To compare designs with one another and with SurrogateOpt on shaping alone, we evaluate every design at the ARIES-CS operating point \citep{najmabadi_aries-cs_2008}, $\aminor^{*} = 1.7\unit{\meter}$ and $\langle B\rangle^{*} = 5.86\unit{\tesla}$. Devices are generated using condition vectors $\yb$ from randomly selected test-set devices with loss fraction $\leq 10\%$. After generation, devices are rescaled to ARIES-CS scale prior to following the evaluation protocol. The rescaling uses two exact invariances of fixed-boundary ideal MHD: the boundary is dilated to $\aminor^{*}$ with $\psiedge$ scaled by the square of the dilation factor, the equilibrium is solved, and $\psiedge$ and the pressure profile are then multiplied by $c = \langle B\rangle^{*}/\langle B\rangle$ and $c^{2}$ before a second solve, which leaves the shape, rotational transform and $\beta$ unchanged and sets the size and field exactly. 

\begin{figure}[tbh!]
    \centering
    \includegraphics[width=.8\textwidth]{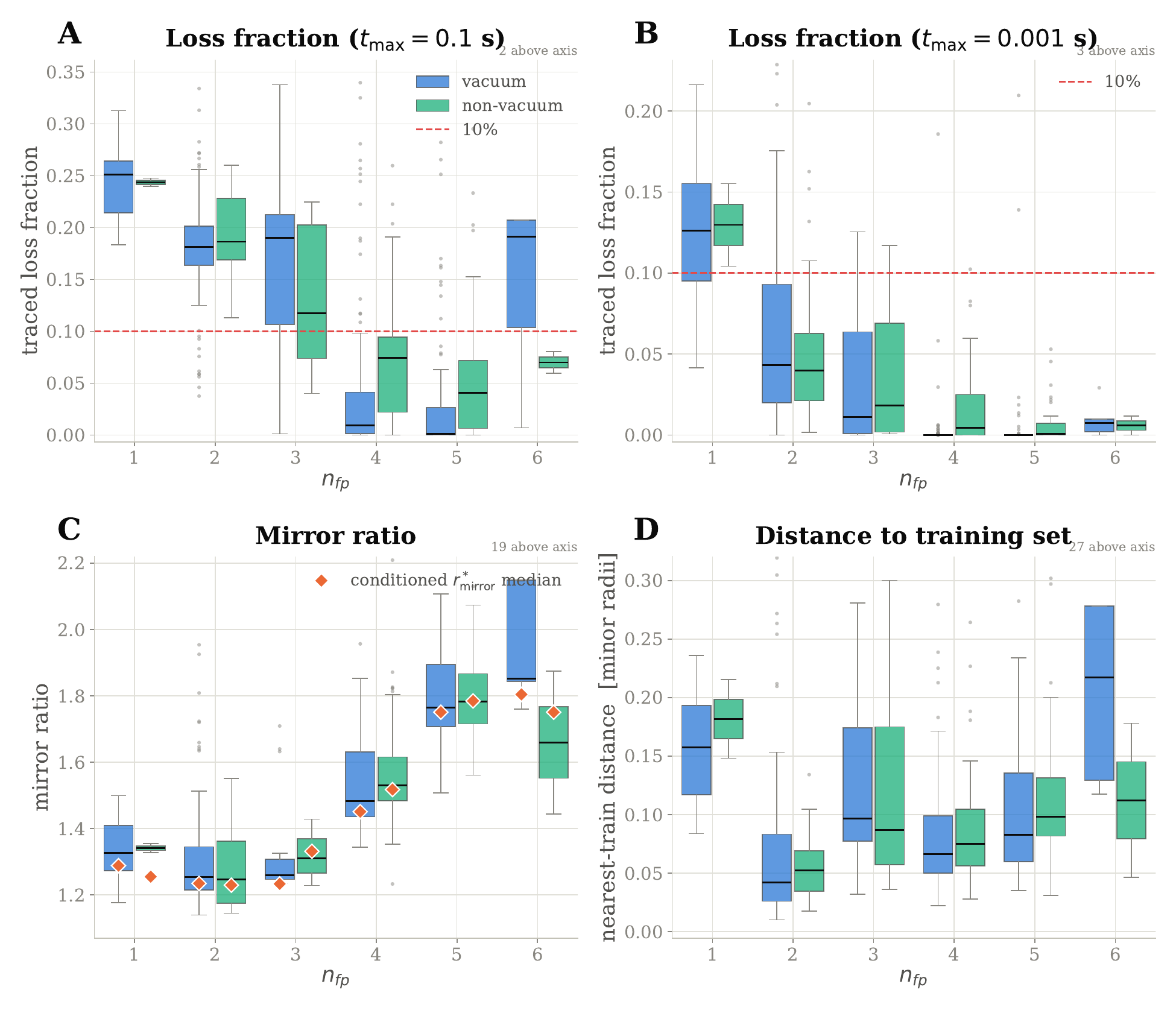}
    \caption{Diffusion at the ARIES-CS scale: the $588$ designs behind the Diffusion column of \Cref{tab:inverse_design_results} (all converged, $586$ traced), by $\nfp$ and vacuum (blue) / finite pressure (green). Panels A and B: traced loss fraction to $\tmax = 0.1\unit{\sec}$ and $1\unit{\milli\second}$. Panel C: distribution of mirror ratio; diamonds indicate the median mirror ratio condition. Panel D: distance to the nearest training boundary (\Cref{app:distance_metric}). The $\nfp = 1$ and $\nfp = 6$ cells hold $23$ and $7$ designs. }
    \label{fig:diffusion_aries_boxplot}
\end{figure}

Of the $588$ designs generated in this fashion, $43.6\%$ of the vacuum and $60.9\%$ of the finite-pressure designs have loss fraction below $10\%$. \Cref{fig:diffusion_aries_boxplot}A and B plot the distributions of loss fraction at $\tmax=0.1\unit{\second}$ and $\tmax=0.001\unit{\second}$ respectively. The distributions are split by the number of field periods, and whether $\beta=0$ (vacuum) or $\beta >0$ (non-vacuum). 
Panel A shows that the low-loss rate is strongly determined by the number of field periods.
Panel B shows that most devices with poor confinement at $\tmax=0.1\unit{\second}$ have good confinement at $\tmax=1\unit{\milli\second}$, except for $\nfp=1$ devices.

Panel C shows the achieved mirror ratios, which track the median conditioned values (diamonds) closely.
Between panels A, B and C it is clear that the model is able to generate devices with low loss fraction and controlled mirror ratios. Most of the low-loss devices, however, have four or five field periods. Lowering the loss target is unlikely to help the two- and three-field-period devices, since the model regresses toward typical losses at low targets (\Cref{appsec:condition_ablation}). Post-training the model is a more promising path to improving model performance 
\citep{domingo2026unified,bergmeister2026reinforce}.

\begin{figure}[tbh!]
    \centering
    \includegraphics[width=\linewidth]{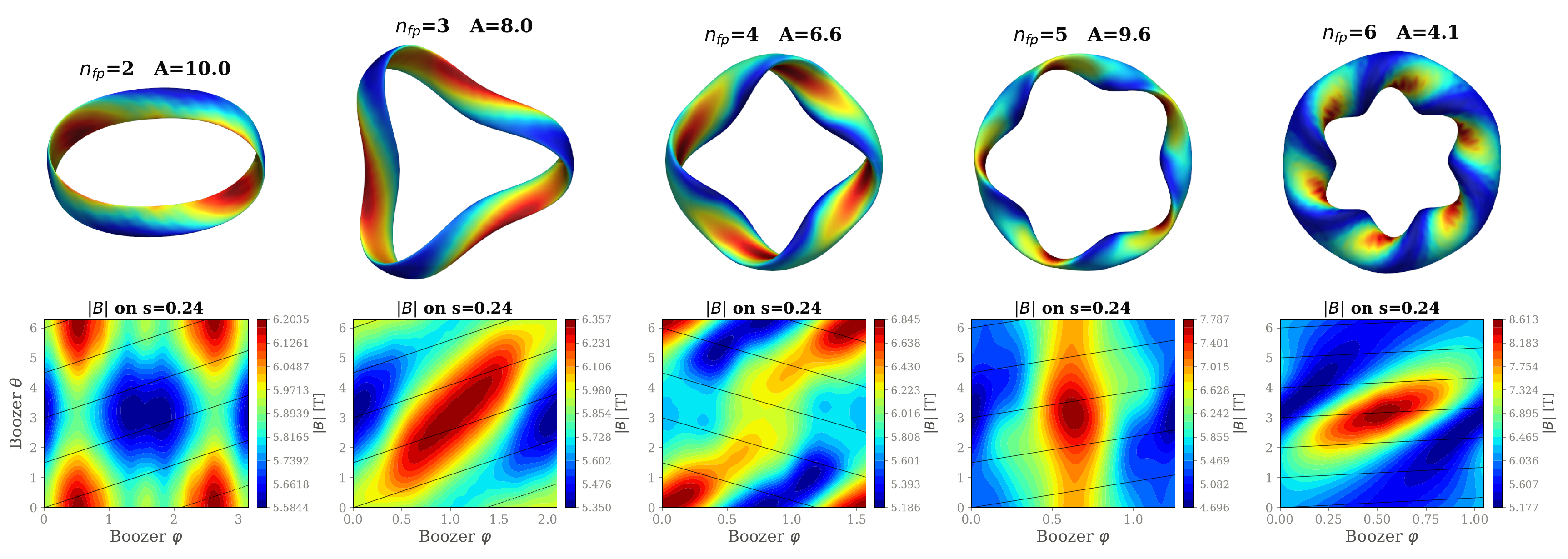}
    \caption{(Top) Five new devices generated by Diffusion, $\nfp = 2$--$6$ from left to right; (bottom) the contours of $|\Bb|$ in Boozer coordinates on $s=0.24$; thin black lines are field lines. The devices have loss fraction $18.5\%, 21.3\%, 8.7\%, 16.3\%, 43.5\%$ at the ARIES-CS scale. The third device ($\nfp = 4$) is a finite-pressure equilibrium ($\beta = 3\%$), the others vacuum.}
    \label{fig:diffusion_novel_devices}
\end{figure}

Panel D plots the distribution of the normalized distance from generated device surfaces to the nearest training surface (\Cref{app:distance_metric}). Since surfaces are normalized to unit minor radius, the distance does not depend on device scale. The generative model finds most of its well-confined designs in the neighbourhood of the training data, but not by copying the devices. 
The designs are never closer than $0.01$ minor radii to a training device ($95\%$ are beyond $0.02$), with a median distance of $0.07$, and often much further.
\Cref{fig:diffusion_novel_devices} exemplifies the variety of devices that can be generated with the model by showing five new devices (top row) and the contours of $|\Bb|$ in Boozer coordinates (bottom row).  
While only one of the five is well confined ($8.7\%$ loss; the others lose $16$--$44\%$), the contours of $|\Bb|$ have unexpected shapes.
None of the devices exhibits precise quasisymmetry (a symmetry of the magnetic field that is sufficient to guarantee good guiding center confinement) \citep{landreman_magnetic_2022}, but the second and third devices from the left could be considered piecewise omnigenous \citep{velasco_piecewise_2024}, indicated by the isolated maxima and minima of $|\Bb|$.

\subsection[Appendix]{Generation with test conditions}
\label{appsec:condition_ablation}

In this section, we generate devices with a \textit{broader} set of conditions and analyze how well designs actually meet their conditions. 

We drew conditions from $5{,}120$ randomly selected test devices, generated one design per condition, and evaluated it (at the size and field strength of the conditioning device, not necessarily ARIES-CS scale); $4{,}420$ converged in \texttt{VMEC} ($13.7\%$ failed) and $4{,}369$ were traced. The set of conditions used here is broader than those used in the ARIES-CS scale generation (\Cref{fig:diffusion_aries_boxplot}) since they are derived from all test devices, not just those with less than $10\%$ losses.

\begin{figure}[tbh!]
    \centering
    \makebox[\textwidth][c]{%
    \includegraphics[width=\textwidth]{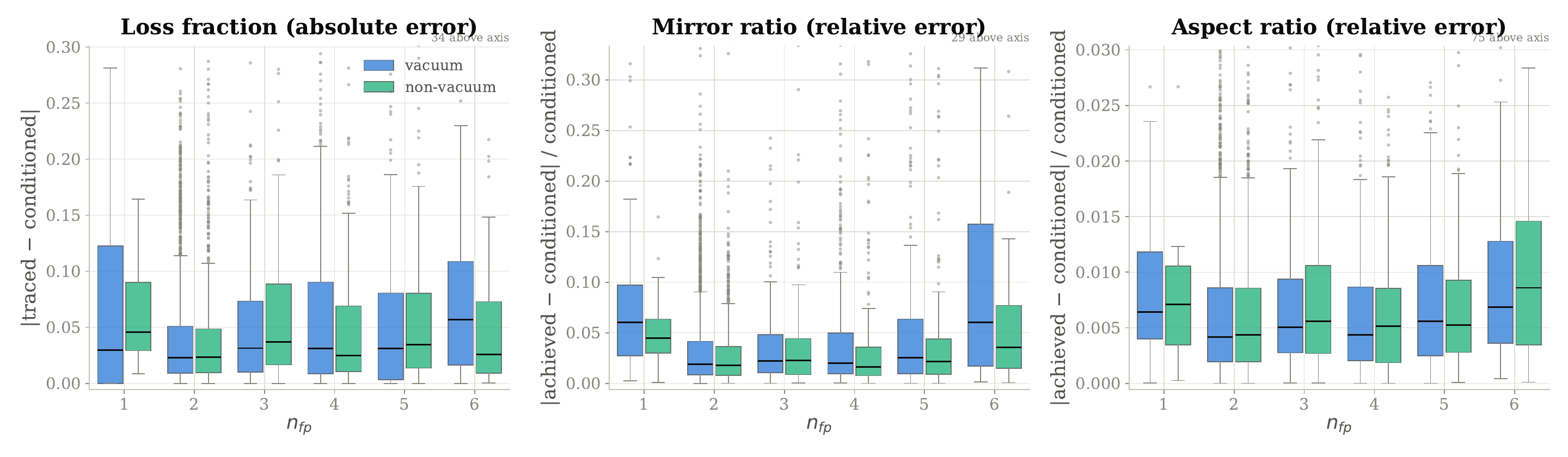}
    }
    \caption{
    Error between achieved and conditioned properties for all $4{,}420$ converged test-conditioned designs at native scale, by $\nfp$ and vacuum (blue) / finite pressure (green): absolute error in loss fraction ($4{,}369$ traced designs), and relative error in mirror ratio and aspect ratio. The vertical scale is set by a few outliers; medians are $0.026$, $2.0\%$ and $0.5\%$. 
    }
    \label{fig:diffusion_errors}
\end{figure}

\Cref{fig:diffusion_errors} shows how closely each of the generated designs meets its conditioning vector. The median absolute error in loss fraction is $0.026$, and the median relative errors are $2.0\%$ in mirror ratio and $0.5\%$ in aspect ratio. The loss-fraction error is not uniform in the target: the median signed error is $+0.085$ for conditioned loss fractions at or below $0.02$, against $+0.003$ for mid-range targets. Thus, at very low-loss targets, the model regresses toward typical devices, and the small overall error partly reflects how many conditioning rows describe devices with moderate---not extremely low---losses. Overall the model is fairly accurate in meeting the test conditions, and this is not at the cost of diversity in designs. Distances to the nearest training surface (using the metric from \Cref{app:distance_metric}) confirm that the sampled surfaces are not memorized training devices: the median test-conditioned design sits $0.126$ minor radii from its nearest training surface, $35\%$ lie beyond $0.2$ and $20\%$ beyond $0.3$, and none of the $4{,}420$ evaluated designs lies within $0.005$ of a training surface.

The generation results depend strongly on the conditions. $7.2\%$ of the generated designs have loss fractions below $10\%$, against the $13.5\%$ rate among the devices whose conditions were used; thus, the model misses some low targets (\Cref{fig:diffusion_errors}). However, when using only conditions from devices with less than $10\%$ losses, $50.0\%$ of the generated designs have loss fractions below $10\%$ ($0 / 11 / 34 / 79 / 84 / 43\%$ at $\nfp = 1$--$6$); this is the same population that is rescaled to the ARIES-CS point above.

\section[Appendix]{Surrogate driven optimization of fast ion losses}
\label{app:loss_optimization}

Surrogate based optimization is used to find stellarator designs with reduced fast ion losses, where neural surrogates are used as predictors for the losses. The optimization variables, $\chib_s$, are the Fourier coefficients $R_{mn}, Z_{mn}$ of a
stellarator-symmetric boundary surface. The objective used, \Cref{eq:opt_problem}, combines the fast ion surrogates with penalties that hold the device at the
requested size, aspect ratio, and field strength.
$m_{\text{energy}}$ is the surrogate prediction of the normalized energy loss at $t_{\max} = 0.1\unit{\sec}$ for
particles launched from $s = 0.1$. The geometric penalties are quadratic about
their targets,
\begin{align}
    J_{\text{aspect}}(\chib) = \frac{\big(A(\chib) - A^*\big)^2}{(A^*)^2},
    \qquad
    J_{\text{minor}}(\chib) = \frac{\big(\aminor(\chib) - \aminor^*\big)^2}{(\aminor^*)^2},\label{eqn:J_aspect}
\end{align}
where $A$ and $a$ are the exact aspect ratio and minor radius of the boundary surface and
$A^*, a^*$ are target values. Discussed in \citep{bindel_direct_2023}, the mirror ratio constraint imposes two one-sided penalties on the
field strength to restrict them to feasible values. Given a target volume-averaged field strength $\langle B\rangle^*$ and target mirror ratio $\rmirror^*$, we define upper and lower bounds on the magnetic field as, $B^*_{\max} = 2 \rmirror^* \langle B\rangle^* / (\rmirror^* + 1)$ and $B^*_{\min} = 2 \langle B \rangle^* / (\rmirror^* + 1)$. The mirror ratio objective is,
\begin{align}
    J_{\text{mirror}}(\chib) = 3 \,\mathcal{E}_{\max}(\chib)^2 + 3\, \mathcal{E}_{\min}(\chib)^2,
    \qquad
    \mathcal{E}_{\max} = \frac{\int_V \max\big(|B| - B^*_{\max},\, 0\big)^2 \, dV}
                              {(\langle B \rangle^*)^2 \int_V dV},\label{eqn:J_mirror}
\end{align}
where the integral is taken over the plasma volume, and $\mathcal{E}_{\min}$ defined analogously for $|B|$ below $B^*_{\min}$. 

Every objective evaluation requires a converged fixed-boundary \texttt{VMEC} equilibrium: $\min|B|$,
$\max|B|$, and the volume weights in $J_{\text{mirror}}$ are computed from it, and the
confinement surrogates. If \texttt{VMEC} fails to converge for a
perturbed boundary the objective returns a fixed moderate value ($10^3$).

Gradients are taken by forward finite differences with an absolute step of $10^{-6}$ on the Fourier
coefficients. The independent \texttt{VMEC} solves are distributed across MPI worker groups, one solve per perturbed coefficient, using \texttt{SIMSOPT}'s
\texttt{MPIFiniteDifference}. The resulting scalar problem is minimized with L-BFGS-B. The surface degrees of freedom are Fourier coefficients up to $\mpol=\ntor=4$, totaling $81$ degrees of freedom. Devices are generated using $\langle B\rangle =5.86, \aminor^* = 1.7$ and $I=0$ and $p(s) = \frac{3}{5}\frac{\langle B\rangle^2  \beta}{\mu_0} (2 - 3s + s^2)$ with $\beta = 0\%, 3\%$. Each run is warm-started from a training device with $A^* \sim U[3.5, 10]$ and run $\nfp \in \{1, \dots, 5\}$. 

\begin{figure}[tbh!]
    \centering
    \makebox[\textwidth][c]{%
    \includegraphics[width=\textwidth]{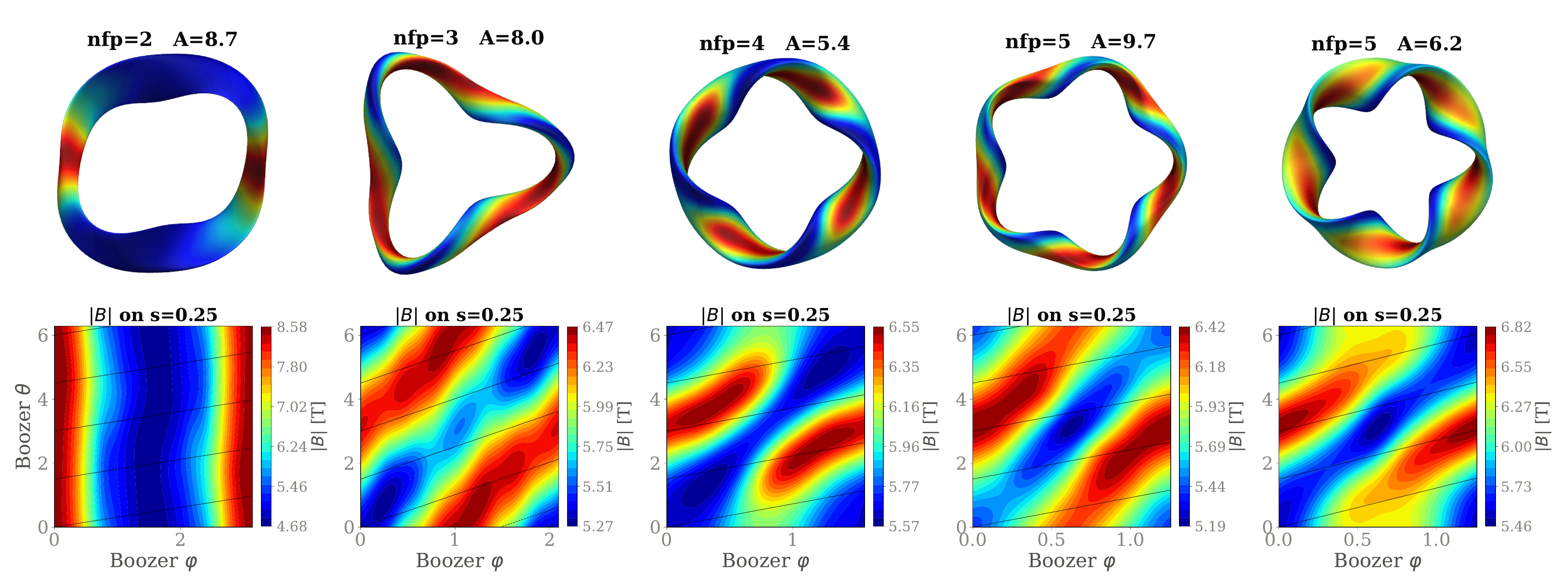}
    }
    \caption{Five stellarators found by surrogate optimization, colored by $|\Bb|$ (top row). (Bottom) Contours of $|\Bb|$ in Boozer coordinates.}
    \label{fig:surrogate_opt_examples}
\end{figure}

\paragraph{Results}
\Cref{fig:surrogate_opt_examples} shows five devices found by the surrogate optimization. The top row shows a top down view of the device while the bottom row shows contours of $|\Bb|$ in Boozer coordinates. Each device is significantly different from every other device in the training set (normalized distance to the training set of at least $0.34$) indicating the ability of the model to generalize. The five devices have loss fractions $5.5\%, 11\%, 3.9\%, 5.8\%, 7.3\%$, mirror ratios less than two. Despite the helical striping shown in the $|\Bb|$, none of the devices are quasisymmetric.

\begin{figure}[tbh!]
    \centering
    \makebox[\textwidth][c]{%
    \includegraphics[width=0.8\textwidth]{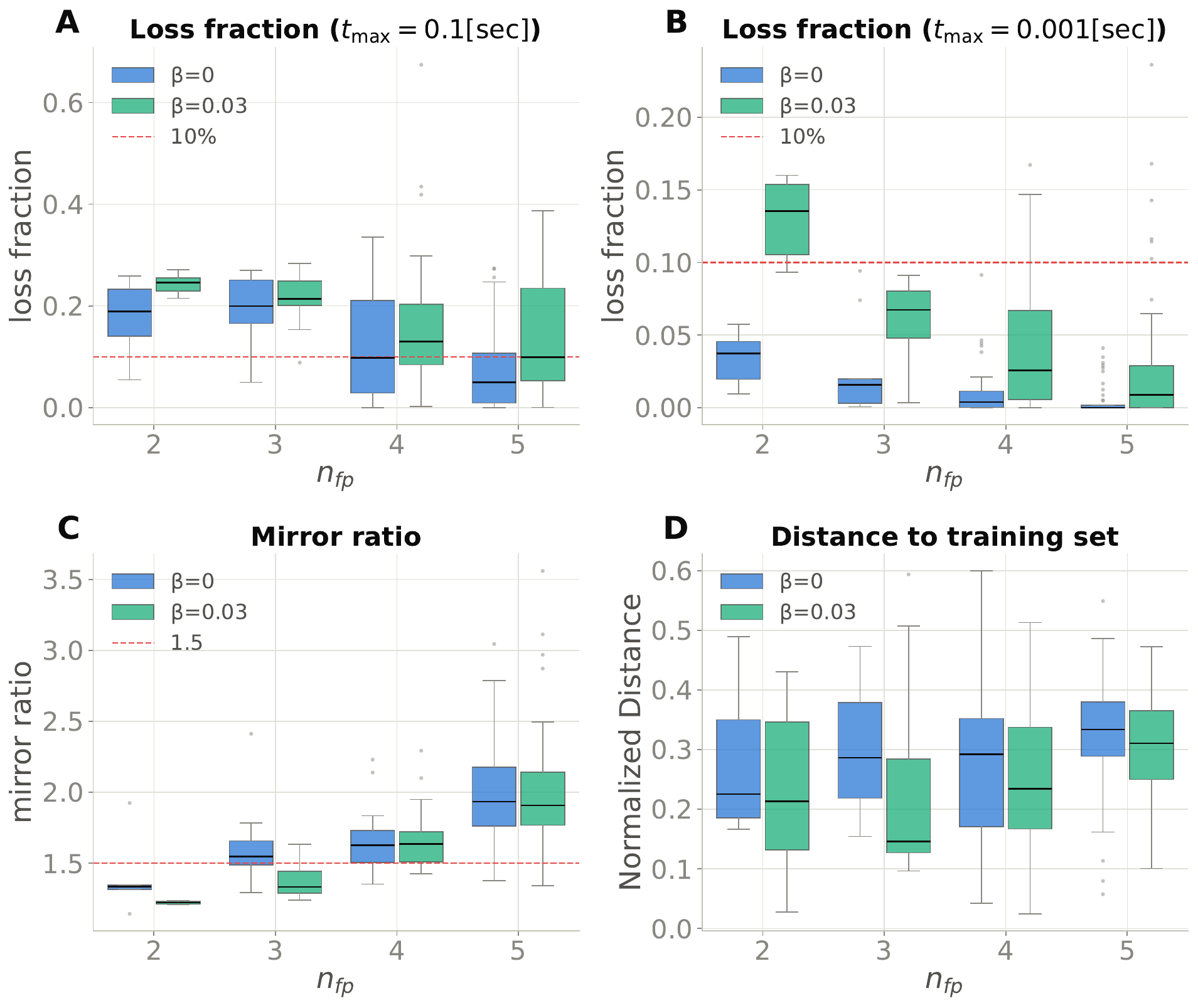}
    }
    \caption{Statistics for 209 solutions of the surrogate optimization problem with predicted loss fraction $\leq 10\%$ to $\tmax = 0.1\unit{\sec}$. Panels A and B show the actual loss fraction to $0.1\unit{\sec}$ and $1\unit{\milli\second}$, showing that half of all solutions for $\nfp=4,5$ have under $10\%$ losses while having reasonable mirror ratios (panel C). The horizontal dashed line indicates target values for each metric. Panel D shows the minimum (normalized) distance between the new surfaces and surfaces in the training set. Most surfaces are fairly distinct, with distance exceeding 20\%(of the minor radius).}
    \label{fig:surrogate_opt_boxplot}
\end{figure}

\Cref{fig:surrogate_opt_boxplot} shows boxplots for the performance of the model at each $\nfp$, for $\beta=0\%,3\%$. Solutions are only included in the box plots if the model for loss fraction predicts less than $10\%$ losses -- more solutions were found that were local minima of the objective with higher loss fractions. Due to the high computational cost of running the experiments, the boxplots consist of only 209 solutions. Panels A and B show the actual loss fraction to $0.1\unit{\sec}$ and $1\unit{\milli\second}$, showing that half of all solutions for $\nfp=4,5$ have under $10\%$ losses while having reasonable mirror ratios (panel C). Panel D shows the distribution of the minimum distance between the new surface and the surfaces in the training set (defined in \cref{app:surfaces}). The optimized surfaces tend to be fairly different from those in the training set, with distance often exceeding 20\%(of the minor radii). 

\subsection[Appendix]{Surrogates for stellarator properties}
\label{app:surrogates_for_properties}

\begin{figure}[tbh!]
    \centering
    \includegraphics[width=\linewidth]{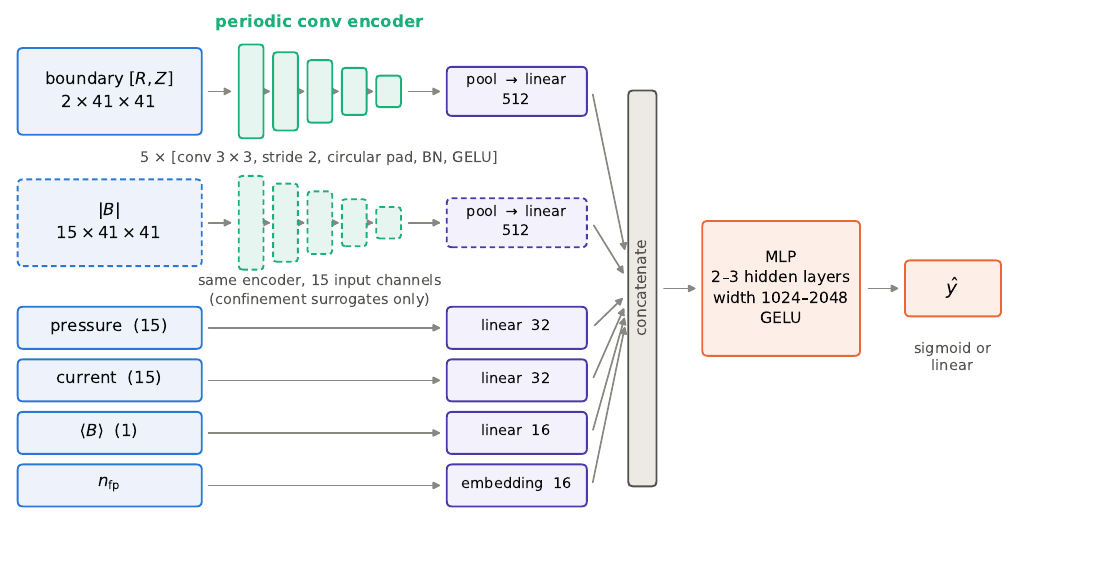}
    \caption{Architecture of the neural network surrogate models for predicting scalar outputs such as loss fraction, energy loss, aspect ratio, etc.}
    \label{fig:surrogate_architecture}
\end{figure}

Each property is predicted by its own scalar-valued network. All of them share the
architecture and training procedure below, differing only in the target, in whether they
take the field strength $|\Bb|$ as an input, and in the width and depth of the final
multilayer perceptron (MLP). $|\Bb|$ is used as an additional input when predicting energy loss or loss fraction, since it generally improves the performance of the predictor. \Cref{fig:surrogate_architecture} shows the model architecture.

The inputs are the plasma boundary, the pressure and net toroidal current profiles, the
volume-averaged field strength $\langle B \rangle$, and the number of field periods
$n_{\mathrm{fp}}$. The boundary is represented pointwise rather than spectrally, as the
cylindrical coordinate fields $R(\phi, \theta)$ and $Z(\phi, \theta)$ on a uniform
$41 \times 41$ grid with $\phi$ spanning one field period, and the profiles are given on
$15$ equally spaced values of the normalized toroidal flux $s$. Since the poloidal angle of
a \texttt{VMEC} boundary is defined only up to an offset and an orientation, every boundary is first
mapped to a canonical parameterization in which $\theta = 0$ lies at the outboard midplane
of the $\phi = 0$ cross section and $\theta$ increases toward $+Z$. This is a single global
phase shift of the $\theta$ grid, so it removes the degeneracy while preserving the Fourier
structure of the surface. Continuous inputs are min--max normalized to $[0, 1]$ using the
training-split range, and $n_{\mathrm{fp}}$ is treated as a categorical variable.

Each input is encoded independently into a fixed-width embedding, the embeddings are
concatenated, and an MLP maps the concatenation to a scalar, as shown in \Cref{fig:surrogate_architecture}. The boundary is encoded as a
two-channel image $[R, Z]$ on the periodic $(\phi, \theta)$ grid by a convolutional network
with circular padding in both angles: five stride-$2$ blocks of a $3 \times 3$ convolution,
batch normalization, and a GELU nonlinearity, with widths $(32, 64, 128, 256, 512)$,
followed by a global average pool and a linear projection to $512$ dimensions. Pooling
rather than flattening makes the embedding invariant to a shift of the sampling grid in
either angle and independent of the grid resolution. When $N$ is used as an input, it is encoded by an
identical convolutional encoding over the same angular grid, with the $15$ flux surfaces entering as input
channels. The profiles are each flattened and linearly projected to $32$ dimensions,
$\langle B \rangle$ to $16$, and $n_{\mathrm{fp}}$ is mapped through a learned embedding
table to $16$. The trunk MLP has $2$--$3$ hidden layers of width $1024$--$2048$ with GELU
activations, depending on which target is being predicted. Targets bounded to $[0, 1]$ by construction, the loss fraction and the
normalized energy, use a sigmoid output so that the prediction respects the bound exactly;
the remaining targets are min--max normalized to $[0, 1]$ and predicted with a linear
output.

Training minimizes the mean squared error in the transformed target with AdamW at a
learning rate of $3 \times 10^{-4}$, weight decay $10^{-4}$, and batch size $1024$. The
learning rate is decayed by a half-cosine schedule to $10^{-6}$, stepped once per
iteration, and gradients are clipped to unit norm. Each model trains for $500$--$700$
epochs on a single NVIDIA A100 GPU.

\section{Heat deposition prediction}
\label{app:heat_prediction}

\paragraph{Estimating Heat Flux}
The heat flux on the plasma boundary is the energy density (energy per unit area) deposited on the boundary by the traced particles: if no particles are lost the heat flux is zero. Suppose $p(\theta, \varphi, E)$ is the distribution of \textit{lost} particles over Boozer coordinates $(\theta, \varphi)$ on the plasma boundary, $s=1$ ($s=0.98$ is used in practice due to numerical reasons discussed in \cref{sec:pipeline}). The heat flux can be computed as 
\begin{align}
    Q_p(\theta, \varphi) = \lossfrac \int E \, p(\theta, \varphi, E) \, dE \quad \quad \left[\frac{\unit{\mega\electronvolt}}{\unit{\radian}^2}\right].
\end{align}
When reporting metrics we divide the field period into an $m\times m$ grid. In the $k$th bin, $\Omega_k$, we compute the mean heat flux as 
\begin{align}
    Q_{p,k} = \frac{1}{\text{Area}(\Omega_k)}\int_{\Omega_k} Q_p(\theta, \varphi) \, d\theta\, d\varphi \quad\quad\left[\frac{\unit{\mega\electronvolt}}{\unit{\meter}^2}\right].
\end{align}
Since our training data and flow matching model provide us with a point cloud of samples of lost particle positions and energies $(\theta_i, \varphi_i, E_i)$ we approximate the losses with Monte Carlo, 
\begin{align}
    Q_{p,k}\approx \frac{1}{n \nfp \text{Area}(\Omega_k)}\sum_{i=1}^{n_{\text{lost}}} E_i\mathbbm{1}\{(\theta_i, \varphi_i\bmod(2\pi/\nfp))\in\Omega_k\} \quad\quad\left[\frac{\unit{\mega\electronvolt}}{\unit{\meter}^2}\right],
\end{align}
where the sum is over the lost particles, $n$ is the total number of particles (lost and confined), and field period symmetry has been exploited to increase the number of samples in the bins.

\begin{table}[tbh!]
\centering
\small
\caption{Evaluation of IID vs PointNet for stellarators from the test set using particles born on $s=0.1,0.25,0.5$. HF and the loss-fraction errors are over the entire test set ($N=74763$); ED needs a lost-particle cloud and is restricted to devices with at least $10$ lost particles ($N=72512$). The subsequent sections show the breakdown of ED and HF when only considering particles born on a given initial surface; or for the subset of configurations within specific loss fraction ranges. Lower is better for every reported metric. \textbf{Bold} = winner per row.}
\label{tab:heat_flux_appendix}
\begin{tabular}{lccc}
\toprule
Metric & IID & PointNet & $\Delta$\% \\
\midrule
\multicolumn{4}{l}{\textit{Headline}} \\
ED (median) & \textbf{0.0306} & 0.0358 & +16.9 \\
\quad MC floor & 0.0193 & 0.0193 & - \\
HF $16\times16$ (median) & \textbf{0.3447} & 0.3824 & +10.9 \\
\quad MC floor & 0.187 & 0.187 & - \\
HF $32\times32$ (median) & \textbf{0.5720} & 0.6134 & +7.2 \\
\quad MC floor & 0.3842 & 0.3842 & - \\
Loss fraction Med-AE & 0.0186 & \textbf{0.0183} & -1.7 \\
\midrule
\multicolumn{4}{l}{\textit{Mean}} \\
ED (mean) & \textbf{0.0425} & 0.0494 & +16.4 \\
\quad MC floor & 0.024 & 0.024 & - \\
HF $16\times16$ (mean) & \textbf{0.5026} & 0.5468 & +8.8 \\
\quad MC floor & 0.185 & 0.185 & - \\
HF $32\times32$ (mean) & \textbf{0.7338} & 0.7809 & +6.4 \\
\quad MC floor & 0.376 & 0.376 & - \\
Loss fraction MAE & 0.0335 & \textbf{0.0324} & -3.4 \\
\midrule
\multicolumn{4}{l}{\textit{ED median by surface}} \\
$s=0.1$ ($N=23520$) & \textbf{0.0334} & 0.0390 & +16.8 \\
$s=0.25$ ($N=24160$) & \textbf{0.0306} & 0.0355 & +16.3 \\
$s=0.5$ ($N=24832$) & \textbf{0.0284} & 0.0333 & +17.2 \\
\midrule
\multicolumn{4}{l}{\textit{HF $32\times32$ median by surface}} \\
$s=0.1$ ($N=24921$) & \textbf{0.5304} & 0.5769 & +8.8 \\
$s=0.25$ ($N=24921$) & \textbf{0.5477} & 0.5896 & +7.6 \\
$s=0.5$ ($N=24921$) & \textbf{0.6325} & 0.6654 & +5.2 \\
\midrule
\multicolumn{4}{l}{\textit{ED median by loss-fraction bin}} \\
$f<0.05$ ($N=2646$) & \textbf{0.1078} & 0.1295 & +20.1 \\
$[0.05, 0.2)$ ($N=6714$) & \textbf{0.0420} & 0.0563 & +34.2 \\
$[0.2, 0.5)$ ($N=32959$) & \textbf{0.0272} & 0.0326 & +19.7 \\
$f \geq 0.5$ ($N=30193$) & \textbf{0.0302} & 0.0337 & +11.7 \\
\midrule
\multicolumn{4}{l}{\textit{HF $32\times32$ median by loss-fraction bin}} \\
$f<0.05$ ($N=4897$) & \textbf{0.0408} & 0.0552 & +35.4 \\
$[0.05, 0.2)$ ($N=6714$) & \textbf{0.2865} & 0.3149 & +9.9 \\
$[0.2, 0.5)$ ($N=32959$) & \textbf{0.4984} & 0.5281 & +6.0 \\
$f \geq 0.5$ ($N=30193$) & \textbf{0.8830} & 0.9809 & +11.1 \\
\bottomrule
\end{tabular}
\end{table}

\paragraph{Results}
\Cref{tab:heat_flux_appendix} shows the performance metrics for the heat flux experiment for each model. The table includes both the headline metrics, as well as a breakdown of energy distance and heat flux by particle birth surface, $s_{\text{init}}$, and over devices binned by loss fraction. The IID model outperforms the PointNet in almost all categories, indicated by the bold entries. The Monte Carlo floor (MC floor) denotes the median value of the metric under sampling noise variation when using the perfect predictor. Almost reaching the MC floor, the IID model demonstrates practical performance.

\subsection{Model architecture and training}
\begin{figure}[tbh!]
    \centering
    \includegraphics[width=\linewidth]{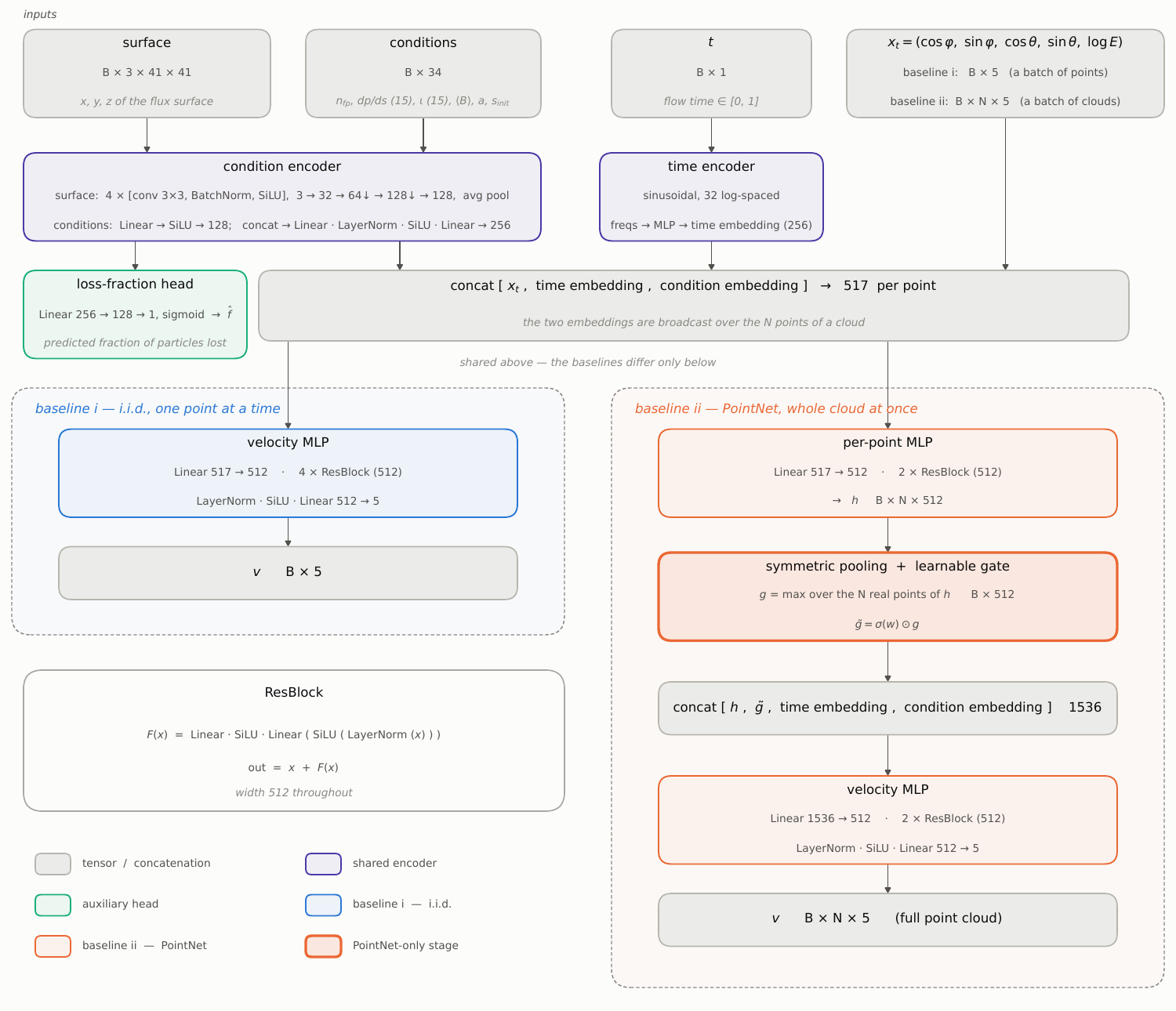}
    \caption{Model architecture for the IID and PointNet velocity models used in the conditional flow matching models for heat flux prediction.}
    \label{fig:heat_models}
\end{figure}

\Cref{fig:heat_models} shows the architecture for the neural network velocity models used within the conditional flow matching methods for heat deposition prediction. The figure shows the shared encoder, followed by the blocks for the two models: the IID velocity model and the PointNet velocity model. We now describe the models training, and sampling procedures.

\paragraph{Inputs and conditioning}
The conditions include the number of field periods, $\frac{dp}{ds}(s)$ and $\iota(s)$ evaluated on $15$ radial surfaces, the
volume-averaged field strength, the minor radius, the surface label the particles are born on $s_{\text{init}}$, and the Cartesian coordinates of its boundary flux surface on a
$41\times41$ grid in the toroidal and poloidal angles,
$\chib_s\in\mathbb{R}^{3\times41\times41}$. A lost particle is represented by its strike location in Boozer coordinates $(\theta, \varphi)$, and energy,
\begin{equation}
  x = \bigl(\cos\varphi,\ \sin\varphi,\ \cos\theta,\ \sin\theta,\ \log E\bigr)\in\mathbb{R}^{5}.
  \label{eq:coords}
\end{equation}
The scalars, the surface coordinates and $\log E$ are standardised using training-set statistics before training.

Both models share one conditioning network. The surface is passed through four
$3\times3$ convolutions with batch normalisation and SiLU activations, of widths
$3\to32\to64\to128\to128$ and stride two in the second and third layer, and then
globally average pooled, which makes the encoder insensitive to the grid
resolution; the scalars are passed through a single affine layer with SiLU
activation. The two representations are concatenated and mapped by a two-layer
network to a 256-dimension embedding. The flow time $t\in[0,1]$
is encoded by a sinusoidal embedding at $32$ frequencies spaced logarithmically over $[1,10^{3}]$ followed by a two-layer network. Every trunk below is a stack of pre-activation residual blocks of width $512$,
\begin{equation}
  \mathcal{R}(z) = z + W_{2}\,\rho\bigl(W_{1}\,\rho(\mathrm{LN}(z))\bigr),
  \label{eq:resblock}
\end{equation}
with $\rho$ the SiLU nonlinearity, $W_1, W_2$ weight matrices, and $\mathrm{LN}$ layer normalisation. The PointNet model is symmetric and permutation equivariant: relabelling the cloud permutes the velocities and changes nothing else.  

Both models are trained by conditional flow matching with a linear interpolant velocity field \citep{lipman_flow_2023}. 
With $x_{1}$ a lost particle of the device $\chib$,
$x_{0}\sim\mathcal{N}(0,I_{5})$, $t\sim\mathcal{U}[0,1]$ and
$x_{t}=(1-t)x_{0}+t\,x_{1}$, the velocity field $v_{\alpha}$ is regressed onto
the constant velocity of the interpolant,
\begin{equation}
  \mathcal{L}_{\mathrm{CFM}}
  = \mathbb{E}\bigl\|\,v_{\alpha}(x_{t},t;\chib_s, \cb) - (x_{1}-x_{0})\,\bigr\|^{2}.
  \label{eq:cfm}
\end{equation}

\paragraph{Generation}
A second model is trained side-by-side, using the same embedding, to predict the loss fraction. The loss fraction model applies a sigmoid to a stack of linear layers with SiLU activations and is trained to minimize the MSE loss. While the flow matching identifies the shape of the point cloud, the loss fraction carries the magnitude of the loss, making distinct devices comparable. When simulating the strike pattern of $n$ points born from some surface, we estimate the loss fraction as $\hat{\lossfrac}$, then sample $n_c = \text{round}(\hat{\lossfrac}n)$ points. The IID model samples points directly from flow matching model one at a time. The PointNet model samples all particles simultaneously, by consuming a whole cloud $x^{1},\dots,x^{n_c}$ and returns one velocity per particle. Particles are generated by integrating $\dot{x}=v_{\alpha}$ from $x(0)\sim\mathcal{N}(0,I_{5})$ to $t=1$ with $50$ uniform explicit Euler steps.

\section{Trajectory prediction}
\label{app:traj_prediction}

This section expands on the discussion of the trajectory prediction task, providing model training details, and more results.

\subsection{Results}
\begin{figure}
    \centering
    \includegraphics[width=\linewidth]{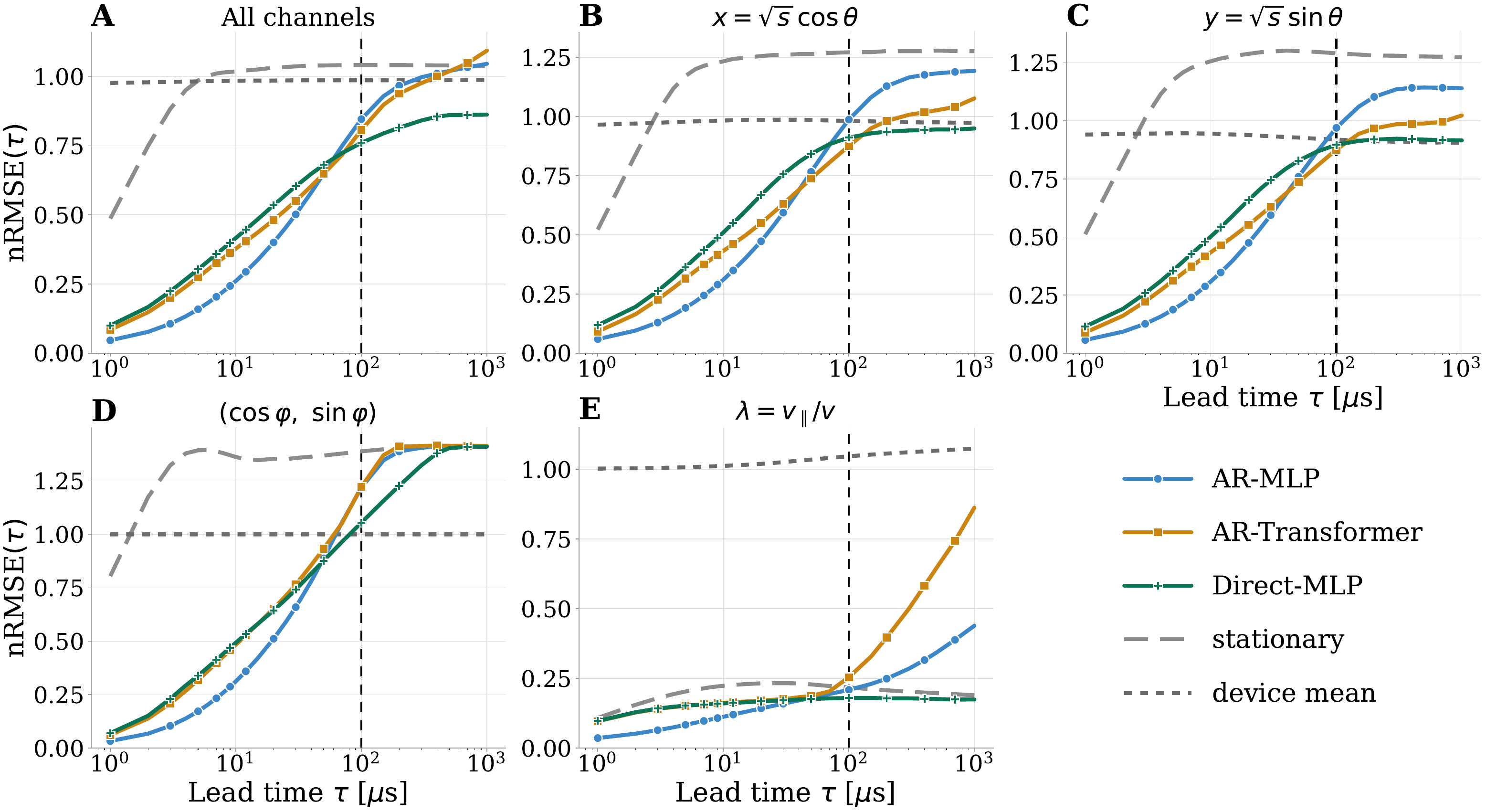}
    \caption{Normalized \textit{test} RMSE across each prediction channel for the trajectory prediction task. The vertical dashed line indicates the training horizon of the autoregressive models (AR-MLP, AR-Transformer). The two baselines ``stationary" and ``device mean" predict the initial condition, and the mean value of channel, respectively, for each device. No model is able to reliably predict any channel to $1\unit{\milli\second}$.}
    \label{fig:traj_pred_metrics}
\end{figure}

\Cref{fig:traj_pred_metrics} shows the \textit{test} nRMSE by channel (panels B-E) over time and the mean value over all channels (panel A). The stationary predictor (predicts $\zb_0$ for all time), and the device-mean (predicts the mean value of each channel for each device) are shown as baselines in the thick and thin dashed lines. A vertical dashed line shows the $100\unit{\micro\second}$ horizon used in training the autoregressive models -- no model performs well beyond this horizon.

\subsection{Models and training}

\paragraph{Output head.}
The particle state at time $t$ is $\zb_t = (\sqrt{s}\cos\theta, \sqrt{s}\cos\theta, \cos\varphi, \sin\varphi, \lambda)$ as defined in \Cref{sec:traj_prediction}. We define three models for predicting the trajectory: AR-MLP, AR-Transformer, Direct-MLP. The ``AR" models are autoregressive while Direct-MLP predicts at arbitrary lead times.
The models consume the current particle state (and in the case of AR-Transformer, the previous $99$ states of its rollout) as well a vector describing the device, $\chib$ (see \Cref{sec:traj_prediction}).
None of the models predict the particle state, $\zb$, directly. Instead, a shared $\Delta$-head, a map to four
increments $(\Delta x, \Delta y, \Delta\varphi, \Delta\lambda)$, updates a reference state:
the poloidal pair, $(x,y) = (\sqrt{s}\cos\theta,\ \sqrt{s}\sin\theta)$, is updated additively to $x\leftarrow x+\Delta x, y\leftarrow y+\Delta y$; the toroidal pair $(\cos\varphi,\sin\varphi)$ is
rotated by $\Delta\varphi$ to $\begin{pmatrix}\cos\varphi \\\sin\varphi\end{pmatrix} \leftarrow \begin{pmatrix}\cos\Delta\varphi & -\sin\Delta\varphi \\ \sin\Delta\varphi & \cos\Delta\varphi\end{pmatrix}\begin{pmatrix}\cos\varphi \\\sin\varphi\end{pmatrix}$; $\lambda \leftarrow \tanh(\operatorname{atanh}\lambda +
\Delta\lambda)$. The toroidal pair therefore stays on the unit circle and $|\lambda|\le1$ holds
exactly, including over long rollouts. $s$ is not capped at 1, so an escaping particle
($s>1$) can be represented. The head is zero-initialized, so every model predicts the identity.

\paragraph{Architectures} 

\begin{figure}[tbh!]
    \centering
    \includegraphics[width=\linewidth]{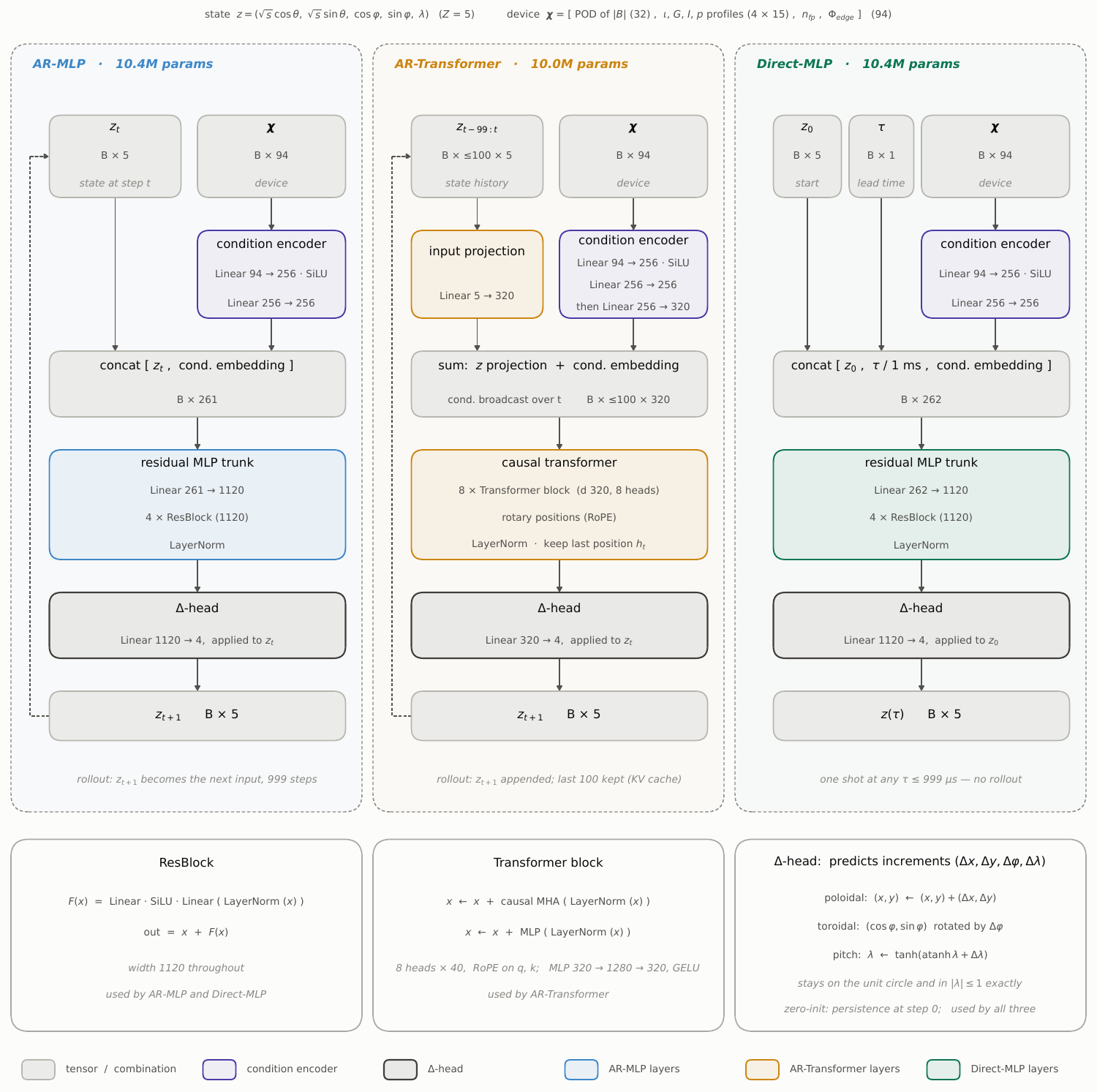}
    \caption{Trajectory model architectures.}
    \label{fig:traj_model_architecture}
\end{figure}
The model architectures are shown in \Cref{fig:traj_model_architecture}. Model sizes are matched, so AR-MLP vs.\ AR-Transformer isolates the
effect of memory, and AR-MLP vs.\ Direct-MLP isolates that of autoregression. Every model embeds $\chib$ with its own two-layer MLP
($94\to256\to256$, SiLU).

AR-MLP is a memoryless one-step map $\hat{\zb}_{t+1} = f(\hat{\zb}_t, \chib)$: $\hat{\zb}_t$ and the condition
embedding is concatenated, projected to width 1120, and passed through four pre-norm residual MLP
blocks and a final LayerNorm (10.4M parameters). 

AR-Transformer adds memory,
$\hat{\zb}_{t+1} = f(\hat{\zb}_{t-(K_{\max}-1):t}, \chib)$: a decoder-only causal transformer with one token per time step
(a linear map $5\to320$, plus the projected condition embedding added at every position), 8 layers,
width 320, 8 heads, a GELU feed-forward of width 1280 and rotary position encodings (RoPE) \citep{su_roformer_2024}. The causal sliding-window mask lets
position $t$ attend only to steps $\max(1,t-(K_{\max}-1)),\dots,t$, so a prediction never sees the future, and never attends further back than the $K_{\max} =100$-step training-horizon: without the window, rollouts beyond the $100\unit{\micro\second}$ would attend to offsets never seen before in training. RoPE rotates queries and keys by angles proportional to their position,
so attention depends only on the relative offset between time steps. The last
position feeds the $\Delta$-head. The
transformer uses a KV cache from which entries older than 100 steps are evicted. 

Direct-MLP predicts $\hat{\zb}(t) = F(\zb_0,t,\chib)$ in one forward pass. It uses the AR-MLP architecture with $t/1\,\mathrm{ms}$ as a sixth input, and its $\Delta$-head
is applied to $\zb_0$ (10.4M).

\paragraph{Loss}
All models share one per-state loss,
\begin{equation}
\ell(\hat \zb, \zb) = \tfrac12\big\|\hat z_{1:4} - z_{1:4}\big\|^2 + (\hat z_5 - z_5)^2,
\end{equation}
where the squared error on the toroidal terms is equivalent to $1-\cos(\hat\varphi-\varphi)$. The autoregressive models are
trained on a $K$-step pushforward loss $\frac1K\sum_{k=1}^{K}\ell(\hat \zb_{t+k}, \zb_{t+k})$, where
gradients propagate through the whole rollout from a true $\zb_t$. $K$ is drawn per batch, log-uniformly on
$\{1,\dots,K_{\max}\}$. The horizon of $100\unit{\micro\second}$ was chosen instead of $1\unit{\milli\second}$ due to the high computational cost of long rollouts. Direct-MLP is trained on $\ell(F(\zb_0,t,\chib), \zb(t))$, with $\zb_0$ the first
sample of the trajectory and $t$ drawn log-uniformly from $1\,\unit{\micro\second}$ up to the trajectory's last
on-grid sample: for lost particles, the final sample is recorded at the moment of loss and so lies
off the $1\,\unit{\micro\second}$ grid. The autoregressive models step on that grid and discard this sample.

\paragraph{Training}
Training uses 471 of the 479 \texttt{train\_split} files (18{,}840 devices). The remaining 8
files are used only to monitor held-out loss. The last 1000 of each device's 6000 trajectories are
never used for training: they form the held-out-particle evaluation set. Samples are streamed
directly from the HDF5 files, drawing 32 samples per trajectory and mixing them through a
$10^6$-sample shuffle buffer so that each batch spans many devices. All runs use AdamW
($\beta = (0.9, 0.95)$, no weight decay), learning rate $3\times10^{-4}$ with a 2000-step linear
warmup and cosine decay to zero, gradient clipping at 1.0, seed 0, and a single
NVIDIA A100/H100 GPU. The reported checkpoint minimises the
2000-step running mean of the training loss. AR-MLP and Direct-MLP train for 400k steps at batch
size 4096 (10.5\,h and 3.5\,h). AR-Transformer trains for 200k steps at batch size 512 (20.7\,h). Its
batch size is limited by memory, because backpropagating through a $K_{\max}$-step rollout keeps every
step's activations even with KV caching. It therefore sees $16\times$ fewer training samples than
AR-MLP.

\paragraph{Ablation over training horizon}

\begin{figure}[tbh!]
    \centering
    \includegraphics[width=\linewidth]{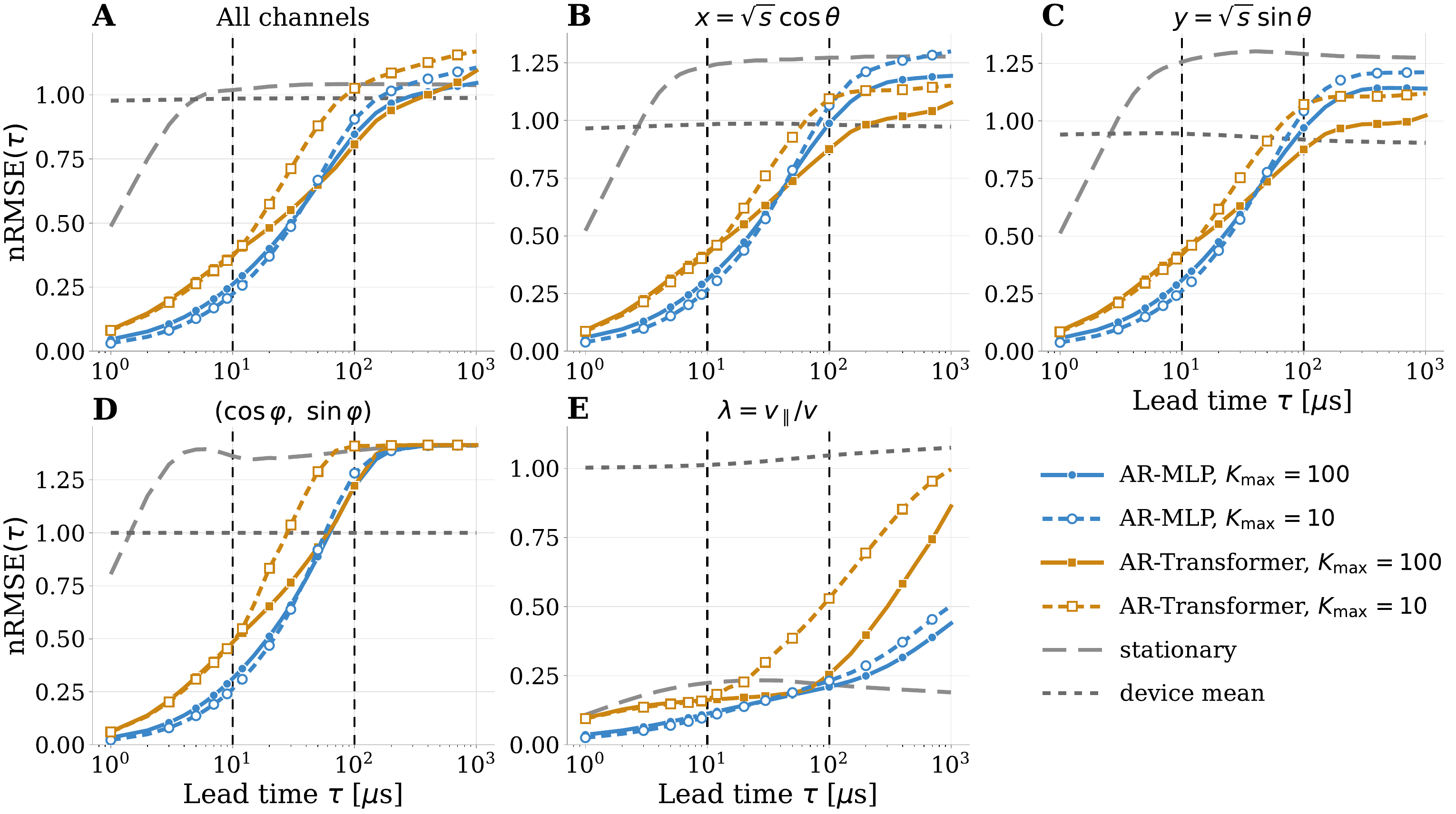}
    \caption{Normalized \textit{test} RMSE across each prediction channel for the trajectory prediction ablation. The vertical dashed lines indicate the training horizons $K_{\max}=10$ and $K_{\max}=100$. The two baselines ``stationary" and ``device mean" predict the initial condition, and the mean value of channel, respectively, for each device. The $K_{\max}=100$ models reliably outperform the $K_{\max}=10$ models beyond a lead time of roughly $30$.}
    \label{fig:traj_pred_metrics_ablation}
\end{figure}

The performance of the autoregressive models is largely dependent on the training horizon, $K_{\max}$: the training loss penalizes errors in predictions up to $K_{\max}$ steps in the future. Here, we ablate over the parameter $K_{\max}$ by showing performance metrics for AR-MLP and AR-Transformer trained on two horizons $K_{\max}=10$ and $100$. Each $K_{\max}=10$ model is trained with the same settings as its $K_{\max}=100$ counterpart (architecture, steps, batch size, learning-rate schedule, and seed); only $K_{\max}$ differs.

\Cref{fig:traj_pred_metrics_ablation} shows the \textit{test} nRMSE by channel (panels B-E) over time and the mean nRMSE over all channels (panel A). The stationary predictor (predicts $\zb_0$ for all time), and the device-mean (predicts the mean value of each channel for each device) are shown as baselines in the thick and thin dashed lines. Vertical dashed lines show the $10\unit{\micro\second}, 100\unit{\micro\second}$ horizons used in training the autoregressive models. Beyond a lead time of roughly $30\unit{\micro\second}$, the $K_{\max}=100$ models reliably outperform the $K_{\max}=10$. While the performance of all models degrades beyond their training horizon, the $K_{\max}=100$ models perform well over a wider range. The difference between the two transformer models, however, is much larger than the difference between the AR-MLP models; memory-based models improve when increasing context length. This evidence suggests that increasing the training horizon to $K_{\max}=1000$ to include the entire trajectory could improve overall performance for the transformer.

\begin{table}[t]
\centering
\caption{Training-horizon ablation, comparing the performance of the autoregressive models trained with $K_{\max}=10$ and $K_{\max}=100$. Lower is better for every metric. \textbf{Bold} = best per row.}
\label{tab:traj_horizon_ablation}
\begin{tabular}{lccccc}
\toprule
Metric & Stationary & \begin{tabular}[b]{@{}c@{}}AR-MLP\\ $K_{\max}=10$\end{tabular} & \begin{tabular}[b]{@{}c@{}}AR-MLP\\ $K_{\max}=100$\end{tabular} & \begin{tabular}[b]{@{}c@{}}AR-Transformer\\ $K_{\max}=10$\end{tabular} & \begin{tabular}[b]{@{}c@{}}AR-Transformer\\ $K_{\max}=100$\end{tabular} \\
\midrule
nRMSE ($\tau = 10\,\mu$s) & 1.019 & \textbf{0.223} & 0.261 & 0.371 & 0.378 \\
nRMSE ($\tau = 100\,\mu$s) & 1.042 & 0.905 & 0.846 & 1.026 & \textbf{0.807} \\
nRMSE ($\tau = 999\,\mu$s) & \textbf{1.038} & 1.107 & 1.046 & 1.170 & 1.094 \\
Loss fraction MAE & 0.183 & 0.133 & \textbf{0.104} & 0.111 & 0.118 \\
Confinement time $W_1$ [$\mu$s] & 159.8 & 91.0 & \textbf{79.4} & 92.0 & 105.2 \\
\bottomrule
\end{tabular}
\end{table}

\Cref{tab:traj_horizon_ablation} compares the key performance metrics for each model with bold indicating the best performance per row. While increasing $K_{\max}$ tends to help AR-Transformer in terms of nRMSE, it does not lead to improvements in loss fraction estimation or confinement time $W_1$. On the other hand, AR-MLP marginally only slightly improved its prediction nRMSE by increasing $K_{\max}$ and also saw a significant improvement to loss fraction and confinement time $W_1$ predictions. Overall, increasing $K_{\max}$ may be beneficial for both model classes, but tuning other hyperparameters may provide greater benefit.

\end{document}